\documentclass[journal]{IEEEtran}
\pdfoutput=1
\usepackage{graphicx}
\usepackage{amsmath}
\usepackage{amssymb}
\usepackage[noadjust]{cite}
\usepackage{float}
\usepackage{color}
\usepackage{url}
\usepackage{dblfloatfix} 
\usepackage{xfrac}
\usepackage{amsfonts}
\usepackage{enumitem}
\usepackage{float}
\usepackage{color}
\usepackage{cite}
\usepackage{graphicx}
\usepackage{epstopdf}
\usepackage{bm}
\usepackage{subcaption}
\usepackage{relsize}
\usepackage{mathtools}
\usepackage{array}
\usepackage{array}
\usepackage{multirow}
\usepackage{multicol}
\usepackage{mathtools}

\usepackage{xcolor}
\usepackage{amssymb}
\usepackage{pifont}
\newcolumntype{P}[1]{>{\centering\arraybackslash}p{#1}}
\everypar{\looseness=-1}
\usepackage{soul}
\graphicspath{{Figures/}}
\begin{document}
\title{Coverage Hole Detection for mmWave Networks:\\ An Unsupervised Learning Approach
} 
\author{Chethan K. Anjinappa and \.{I}smail G\"{u}ven\c{c},~\IEEEmembership{Fellow, IEEE} \\
Department of Electrical and Computer Engineering, NC State University, Raleigh, NC 27606\\
e-mail:~\{canjina,iguvenc\}@ncsu.edu  
\thanks{This work has been supported in part by NSF through CNS-1916766 and CNS-1939334, and by DOCOMO Innovations, Inc. We would like to thank Nadisanka Rupasinghe and Haralabos Papadopoulos from DOCOMO Innovations, Inc. for valuable advice and insightful conversations.
}
\vspace{-.5cm}
}

\renewcommand{\baselinestretch}{.890}

\maketitle
\begin{abstract}
The utilization of millimeter-wave (mmWave) bands in 5G networks poses new challenges to network planning. Vulnerability to blockages at mmWave bands can cause coverage holes (CHs) in the radio environment, leading to radio link failure when a user enters these CHs. Detection of the CHs carries critical importance so that necessary remedies can be introduced to improve coverage. In this letter, we propose a novel approach to identify the CHs in an unsupervised fashion using a state-of-the-art manifold learning technique: uniform manifold approximation and projection. The key idea is to preserve the local-connectedness structure inherent in the collected unlabelled channel samples, such that the CHs from the service area are detectable. Our results on the DeepMIMO dataset scenario demonstrate that the proposed method can learn the structure within the data samples and provide visual holes in the low-dimensional embedding while preserving the CH boundaries. Once the CH boundary is determined in the low-dimensional embedding, channel-based localization techniques can be applied to these samples to obtain the geographical boundaries of the CHs.\looseness = -1
\end{abstract}

\begin{IEEEkeywords}
5G, coverage hole, dimensionality reduction, mmWave, radio hole, t-SNE, UMAP, unsupervised learning.
\end{IEEEkeywords}

\vspace{-.4cm}
\section{Introduction}
\IEEEPARstart{F}{ifth}-generation~(5G) networks utilize millimeter-wave~(mmWave) bands to deliver high data rates with low latency, exceeding what is conceivable with the traditional sub-6 GHz cellular systems. 
Due to their unique characteristics, mmWave signals are vulnerable to blockages, creating more \emph{coverage holes} (CH) in a radio environment than sub-6~GHz bands. These CH regions can formally be defined as the area of radio coverage that is below a {threshold level of received signal strength (RSS)} needed for robust radio performance and is often caused by physical obstacles such as new buildings and grown-up trees~\cite{gomez2016method}. When the users enter these CHs, they will suffer from call drops and radio link failures. Mitigation of these CHs is often taken care of diligently during the network-planning. Despite careful network-coverage planning~\cite{anjinappa2020base}, anomalies such as CHs may still exist due to the dynamic changes in the environment. Thus, it is necessary to detect CH boundaries and then deploy remedies such as careful placement of intelligent meta-surfaces/passive metallic reflectors~\cite{anjinappa2020base} to mitigate~CHs.\looseness = -1


The traditional way to detect CHs in legacy cellular networks is to rely on a combination of drive tests, customer complaints, and software/hardware alarms~\cite{akbari2016reliable}. These techniques are not only time-consuming but are also costly and unreliable. In an endeavor to overcome these challenges, the minimization of drive test (MDT) technique has been standardized by 3GPP~\cite{hapsari2012minimization}. The MDT allows the serving base station~(BS) to leverage the measurement reports (includes RSS, geographical location) from user equipments (UEs) to generate coverage maps that help to capture CHs. The MDT may still suffer from positioning/quantization errors and scarcity of UE reports. Moreover, these methods extensively rely on the UE location which also raises the concern of intrusion of location privacy~\cite{grissa2017location}. If the location services are turned off then the UE locations should be estimated by localization techniques such as~\cite{ruble2018wireless}, which might yield erroneous coverage~maps.

Conversely, current cellular networks under-utilize a large amount of channel information that is put into practice for different network operations. {For instance, the channel estimated based on pilot transmission is used within a particular frame (e.g., for precoding and data detection) and discarded {afterward} without being further utilized.} These estimates could be stored in a historical database and be utilized for learning anomalies such as CHs with no extra-cost of measurement resources. We assume only the channel estimates are stored, and the associated spatial locations are unknown, viewed as unlabelled data samples. The availability of such large quantities of unlabelled data at the BS would enable end-to-end data-driven unsupervised {machine learning~(ML) methods to learn the hidden spatial structure embedded within the data.} 

In this letter, we detect CHs in an unsupervised fashion leveraging {the state-of-the-art} data-driven unsupervised learning algorithm named \textit{uniform manifold approximation and projection} (UMAP)~\cite{mcinnes2018umap}. {We do so by generating a low-dimensional embedding (dimension of 2 for visualization) from the high-dimensional channel state information available at the BS (input dimension depends on representation type of the channel samples)}. In particular, we seek to preserve the pairwise distance structure amongst all the data samples, a balance of local-global structure preservation, which would help us to detect CHs. The idea of neighborhood preservation (local-connectedness) is investigated in~\cite{studer2018channel} which motivated our present work. Our results demonstrate that the proposed method can learn the inherent structure within the data and provides visual holes while maintaining CH boundaries in the low-dimensional embedding. To the best of our knowledge, this is the first attempt to use unsupervised learning for detecting CHs without using the location information or measurement reports. {For qualitative comparison, we compare against the popular linear and state-of-the-art manifold learning dimensionality reduction techniques used for visualization.}\looseness = -1


\section{System Model}\label{Sec:Sys_Model}

\subsection{System Model and Data Available at the BS}
Consider an urban mmWave outdoor-to-outdoor communication scenario, where a BS is serving the outdoor ground~UEs in a given service area~(SA). {Formally, the SA is defined as the set of 
$(x, y, z)$ coordinates that are served by the BS. Let $\boldsymbol{\ell}_m = [x_m, y_m, z_m]^{{\text{T}}} \in \mathbb{R}^3$ hold the 3D-coordinates of {$m$-{th}} spatial location out of the $M$ distinct spatial locations in the~SA. We denote the collection of the $M$ spatial locations by}\looseness = -1
\begin{equation}
\mathcal{L} = \{ \boldsymbol{\ell}_1, \boldsymbol{\ell}_2, \ldots , \boldsymbol{\ell}_M \}~,
\end{equation}
For simplicity, we assume the BS is equipped with a uniform linear array (ULA) of {$N$} antennas, and the UEs have a single antenna, respectively. In a static scenario, the frequency-domain wideband channel $\mathbf{h}_m \in \mathcal{C}^{{N} \times 1}$ between the BS and the UE at $\boldsymbol{\ell}_m$ is expressed as:\looseness = -1
\begin{equation}\label{Eq:Channel}
    \mathbf{h}_m(f) = \sum_{k=1}^{K} \alpha_{k,m} \mathbf{a}(\theta_{k,m}) \exp(-j2\pi \tau_{k,m} f)~,
\end{equation}
where the channel is characterized by $K$ multi-path components (MPCs), with $\alpha_{k,m}, \theta_{k,m}$, and $\tau_{k,m}$ being the complex amplitude, azimuthal angle-of-arrival~(AoA), and delay of the $k^\text{th}$ propagation path for the UE at $\boldsymbol{\ell}_m$, respectively. The term $\mathbf{a}(\theta)$ denotes the normalized BS array response, and the half-wavelength spacing array response is expressed as $\mathbf{a}(\theta) =  \frac{1}{\sqrt{{N}}}[1 \exp(j\pi\cos(\theta)) \ldots \exp(j\pi({N} -1)\cos(\theta))]^\text{T}$. Clearly, the structure of the channel $\mathbf{h}_m$ in \eqref{Eq:Channel} reveals that it is characterized by the propagation paths which in turn depends on the spatial locations $\boldsymbol \ell_{m}$ and other geometric features of the propagation environment. {This can be best thought of as an existence of non-linear mapping function from the spatial locations to the corresponding channel samples and vice-versa.}

For the CH detection, we assume that the channel information from all the distinct spatial locations is available at the BS without their corresponding location information. This collection of channel samples is denoted by
\begin{equation}\label{Eq:Channel_Set}
    \mathcal{H} = \{ \mathbf{h}_1 , \mathbf{h}_2 , \ldots , \mathbf{h}_M\}~.
\end{equation}
Note that in a practical system, the UEs send pilot symbols~$\mathbf{s}$ and the BS estimates the channel $\mathbf{h}$ based on the noisy measurements, $\mathbf{y} = \mathbf{h} \mathbf{s} + \mathbf{n}$, where $\mathbf{n}$ is {the additive white Gaussian noise.} Instead of discarding these channel estimates, we assume that the BS collects them in vast quantities to form~$\mathcal{H}$. A similar assumption is made in studies such as~\cite{studer2018channel}.\looseness = -1 

With these in perspective, our main goal in this  paper is {``\textit{to detect and locate the CHs, if there is any, using the collected channel samples, $\mathcal{H}$, and in particular, without considering the spatial location information}."} Above is a classical unsupervised ML task where the aim is to learn how to best use the hidden structure within the data. We learn this up next.\looseness = -1

\subsection{Coverage Hole Formation}
\begin{figure}[t]
\centerline{\includegraphics[scale=0.105]{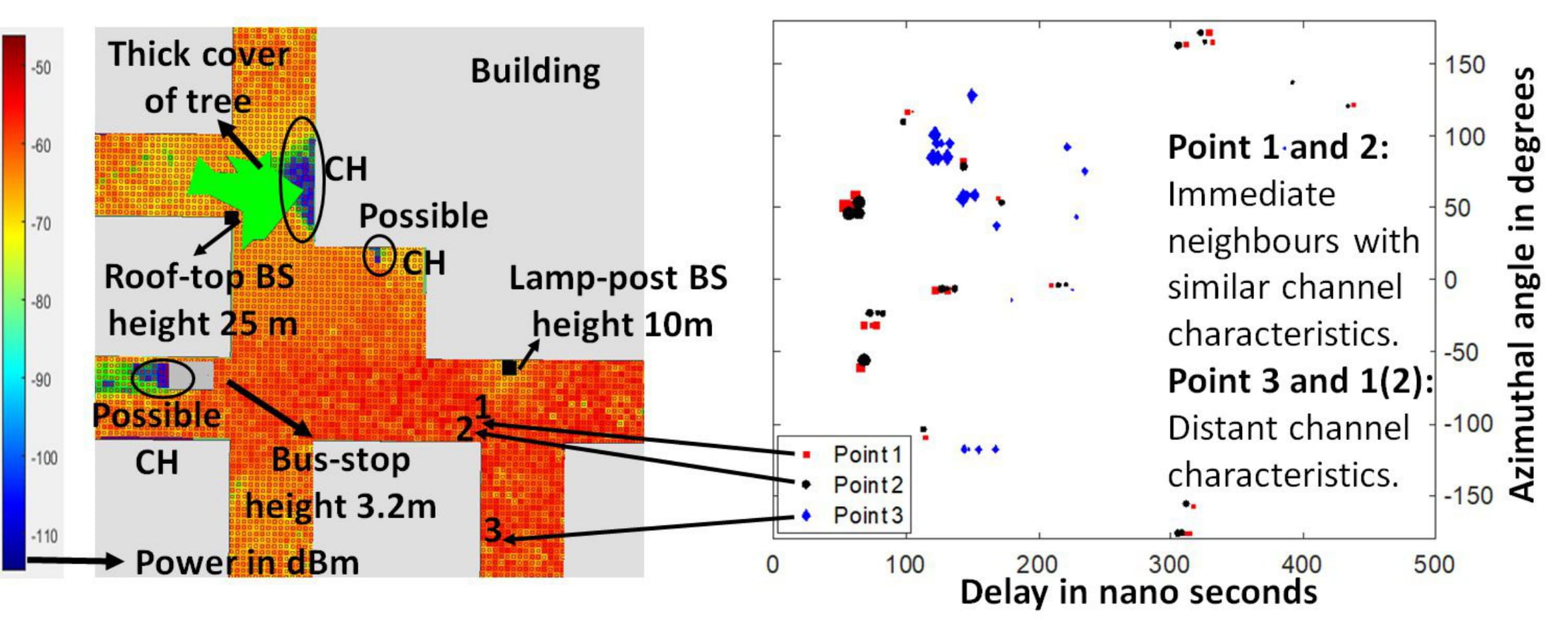}}
\caption[ ]{An example RT scenario for CH illustration. The CH may happen in numerous ways, such as a newly built bus-stop or a thickly spread tree\footnotemark.}
\label{Fig:CH_Illustate}
\vspace{.1cm}
\end{figure}

Formally, as defined before, CH is the area of radio coverage that is below a {threshold level} needed for robust radio performance. An illustration of a CH through a simple ray-tracing (RT) scenario is shown in Fig.~\ref{Fig:CH_Illustate}. {The prime cause of a CH is the lack of significantly powered (or no) MPCs reaching that point (due to the geometry of the environment), resulting in RSS lower than the required threshold. Whereas, outside of the CH, there exist MPCs with sufficient power resulting in RSS suitable for robust radio connection}, as can be seen in Fig.~\ref{Fig:CH_Illustate}. {Mathematically, the CH region $\mathcal{L}_\text{CH}$ can be represented as $\mathcal{L}_\text{CH} = \{ \boldsymbol{\ell}_m | RSS(\boldsymbol{\ell}_m ) < RSS_0,~ \forall m \in [M]\}$, and it can be a null set if the CHs do not exist.} Thus, there will be a drastic change in the propagation characteristics across the CH boundary. These propagation characteristics will further be captured in a non-linear fashion by the high-dimensional channel matrix, as in~\eqref{Eq:Channel}.\looseness = -1

{The above argument suggests that, outside the CH, if two UEs are near each other (neighbors), then the channels at these points are somewhat similar (locally-connected) in the high-dimensional space. This is due to the spatial consistency, as the nearby points should see similar scatterers and have comparable propagation characteristics, as illustrated in Fig.~\ref{Fig:CH_Illustate}.} On the other hand, UEs located far-away will have different propagation characteristics resulting in distinct channels, signifying that the points are not neighbors. A similar argument applies to the data samples inside the CH. {However, at the boundary of the CH, the channels will be distinct even for neighboring UEs as there is an abrupt change in the propagation characteristics - a key characteristic leveraged to detect~CHs.}\looseness = -1 %

\footnotetext{Black Gum/Northern red oak trees used in American urban areas can spread out up to 40-45 feet, multiple such trees on a roadside can block the street leading to possible CHs from the BSs.\looseness = -1}

To exploit the above structure, it is thus desirable to have an algorithm that does the following two things: 1) {capturing} the pairwise distance structure amongst all the data samples (local-global distance), and 2) {preserving} the captured connected-structure in the low-dimensional representation. We hypothesize that this strong tight-preservation of local-global geometry in the lower dimension would enforce to have a visual representation of the CH. {In some sense, what we describe is an approach to connect similar points tightly, and the CH will consequently be recognized as a visual hole/discontinuity within the low-dimensional space}.\looseness = -1



\section{Coverage Hole(s) Detection via UMAP}\label{Sec:UMAP}
We now investigate the core goal of CH detection with the help of UMAP, which is a state-of-the-art dimensionality reduction algorithm. The justification for selecting UMAP over other algorithms is that its characteristics are in harmony with our goal of preserving the local-global distance super-tightly in the low-dimensional embedding.\looseness = -1

\subsection{UMAP: Input, Output, and Hyper-Parameters}
UMAP is a new non-linear dimensionality reduction algorithm proposed by McInnes \emph{et al.}~\cite{mcinnes2018umap}. It seeks to learn the manifold structure in the high dimensional input data and tries to preserve a similar structure in the low dimensional output space. It does so by optimizing the layout of the low-dimensional data space such that the cross-entropy of the structure between the high and low dimensional space is minimized. We now discuss the hyper-parameters that are pivotal in understanding how the UMAP works.

We denote $\mathcal{X} = \{X_1, X_2 , \ldots, X_M \}$ as the high-dimensional input to UMAP and $\mathcal{Y} = \{Y_1 , Y_2, \ldots , Y_M \}	\subseteq \mathbb{R}^{dim}$ as the low-dimensional output of dimension $dim$. The critical hyper-parameters to the algorithm are: 1) (dis)similarity measure metric $d$; 2) neighborhood size $n$; and 3) minimum distance $d_{\rm min}$. The parameter $d$ quantifies the (dis)similarity between any two samples, and we choose the Euclidean distance as the metric in this work. The neighborhood size $n$ considers the number of neighbors to take into account for a local metric approximation of~$d$. The choice of neighbors is a trade-off between capturing the local versus global structure. Smaller values will ensure the local-structure is accurately captured at a loss of global-structure, while the larger values of $n$ will have a vice-versa effect. Finally, $d_{\rm min}$ is an aesthetic parameter that controls the minimum distance that needs to be maintained between the points in the output space. This parameter avoids the potential overlapping of the points that might happen at the output space, which happens to be the case e.g. with the t-SNE technique (see results). Low values of $d_{\rm min}$ ensure densely packed output space, whereas higher values will lead to loosely packed output~space.\looseness = -1

\subsection{UMAP: Working Principle}
With the above things in perspective, we now define the working principle of the UMAP. We emphasize that UMAP is built based on the deep theoretical foundations of topological data analysis and manifold theory. Unfortunately, these details are elided here for the sake of brevity; interested readers may consult~\cite{mcinnes2018umap}. For completeness, we still provide a high-level working principle here. Mathematically, UMAP has two significant steps: 1) weighted neighbor graph construction, which captures the locally-connected structure within the data;  followed by 2) low-dimensional construction, which ensures the above-captured structure is preserved as much as possible. These steps are detailed as follows.\looseness = -1


$1)$ \textbf{Graph Construction:} For a given hyper-parameters setting, compute $n$ nearest neighbors to each input data sample $X_i$ denoted as $d(X_i,X_{i_j}), 1 \leq j \leq n$. This step is similar to the $k$-nearest neighbor algorithm. Further, for each data sample, variables $\rho_i$ and $\sigma_i$ are picked such that:
\begin{eqnarray}
    \rho_i = \min\Big\{ d(X_i,X_{i_j})\big| 1 \leq j \leq n, d(X_i,X_{i_j}) > 0\Big\},\\
    \sum_{j=1}^{n} {\exp \left(\frac{\max\{0, d(X_i,X_{i_j}) - \rho_i\}}{\sigma_i}\right)} = \log_2(n).
\end{eqnarray}
The term $\rho_i$ is the distance to the first nearest neighbor and its contribution to {the exponential kernel weight $w^\text{I}_{ij} = {\exp \left(\frac{\max\{0, d(X_i,X_{i_j}) - \rho_i\}}{\sigma_i}\right)}$ helps to ensure local connectivity.} Due to the construction, it also ensures at least one of the points is connected. The normalization factor~$\sigma_i$, chosen to satisfy the constraint, ensures weights add up to~$\log_2(n)$. \looseness = -1

With this defined, a weighted directed graph $G = (V, E, w^\text{I})$ is constructed where vertices $V$ of the graph are the data samples, edges are $E = \{ (X_i,X_{ij}) | 1 \leq i \leq N, 1 \leq j \leq n \}$, and weights are ${w^\text{I}_{ij}}$. These weights collected via an adjacency matrix signify how much connectivity should be maintained. In our case we use the Euclidean norm as the dissimilarity metric, and hence the created graph would be symmetric. Otherwise, asymmetric weights are handled by the UMAP implementation exploiting the symmetry property.


$2)$ \textbf{Graph Layout (Low-dimensional Representation)}: The final step of low-dimensional embedding construction is taken care of by minimizing the cross-entropy of similarities between the input and output space~as
\begin{equation*}
    C(\mathcal{X},\mathcal{Y}) =  \sum_{i,j} -w^\text{I}_{ij} \log\left( {w^\text{O}_{ij}}\right) - (1-w^\text{I}_{ij}) \log\left( {1 - w^\text{O}_{ij}}\right) ,
\end{equation*}
where, $w^\text{O}_{ij} = (1 + a ||Y_i - Y_j ||_2^{2b})^{-1}$ is the weight term that captures the similarity in the output space between the $i^\text{th}$ and $j^\text{th}$ sample. The scaling parameters $a$ and $b$ are chosen by fitting a non-linear least-square function relating to $d_{\rm min}$. Please consult~\cite[Definition 11]{mcinnes2018umap} for technical details. The key idea here is to place the samples in the low dimensional space such that similarities captured by the objective $C$ is minimized. We emphasize that, in this work, we do not claim the novelty in the algorithm development. However, we claim it for the CH application utilizing an algorithm such as UMAP tailor-made for such applications. Up next, we talk about the features fed to the UMAP.\looseness = -1

\vspace{-.1cm}
\subsection{Features}
We focused on two sets of features: 1) Element-space $\mathcal{H}$, a set of physical channel representation as discussed in \eqref{Eq:Channel}, and 2) virtual-space $\mathcal{H}^\text{V}$, a collection of virtual channel representation of the physical channels \eqref{Eq:Channel_Set}, $\mathcal{H}^\text{V} = \{ \mathbf{h}_1^\text{V} , \mathbf{h}_2^\text{V}, \ldots , \mathbf{h}_M^\text{V}\}$. Here, $\mathbf{h}^\text{V} = {\rm vec}(\mathbf{H}^\text{V})$ is the vectorized virtual representation of the channel~\cite{Virtual_Rep} and is related to the physical channel~as 
\begin{eqnarray}
\begin{aligned}
\mathbf{H}^\text{V}(\theta_\text{D},\tau_\text{D}) 
& = \sum_{k=1}^{K} \alpha_{k} \boldsymbol{f}_{N_\text{D}}(\theta_\text{D} - \theta_k) {\rm sinc}\big(W(\tau_\text{D} - \tau_k )\big)~, 
\end{aligned}
\end{eqnarray}
where $\mathbf{H}^\text{V}$ represents the 2D response in the Fourier angle-delay space over the bandwidth $W$ evaluated at the discretized angle-delay $(\theta_\text{D},\tau_\text{D})$. Further, the term~${\rm sinc}(x) = \frac{\sin(\pi x)}{\pi x}$ and $\boldsymbol{f}_N(x) = \frac{\sin(\pi N x)}{N \sin(\pi x)}\exp(-j2\pi x)$ represent the sinc and normalized-Dirichlet kernel structure, respectively. The angle-delay space is discretized into $N_\text{D}$ and $T_\text{D}$ grids, respectively. Further, both the element-space and the virtual space features are normalized such that each sample {has a unit norm.}\looseness = -1

The above-discussed normalized features are complex numbers in nature. Thus, a non-linear transform of the above features such as real $\mathbb{R}(\cdot)$, imaginary $\mathbb{I}(\cdot)$, angle $\angle(\cdot)$, and absolute $|\cdot|$ parts were passed to the UMAP algorithm as the input. Simulation evidence showed that the absolute-valued normalized-virtual representation $\mathcal{X} = |\mathcal{H}^\text{V}|$ yielded the best result; a similar observation was made in~\cite{studer2018channel}. The possible explanation for this phenomenon is due to the inherent Dirichlet and Sinc structure in the virtual representation that captures the energy according to the propagation characteristics and helps to preserve the neighborhood geometry. Thus, we restrict our discussion and results to the absolute-valued virtual channel.\looseness = -1

\section{Simulation Results}\label{Sec:Results}
In this section, we demonstrate the ability of UMAP to detect CHs with the DeepMIMO dataset~\cite{Alkhateeb2019}. The DeepMIMO channels are constructed based on accurate RT data that captures the dependence on the environment geometry/materials and BS/UE locations. Therefore, the data obtained from the DeepMIMO support spatial consistency and presumed to represent the ground truth close to a realistic environment.\looseness = -1   

\begin{figure}[t]
\centering
{\includegraphics[width=.225\textwidth]{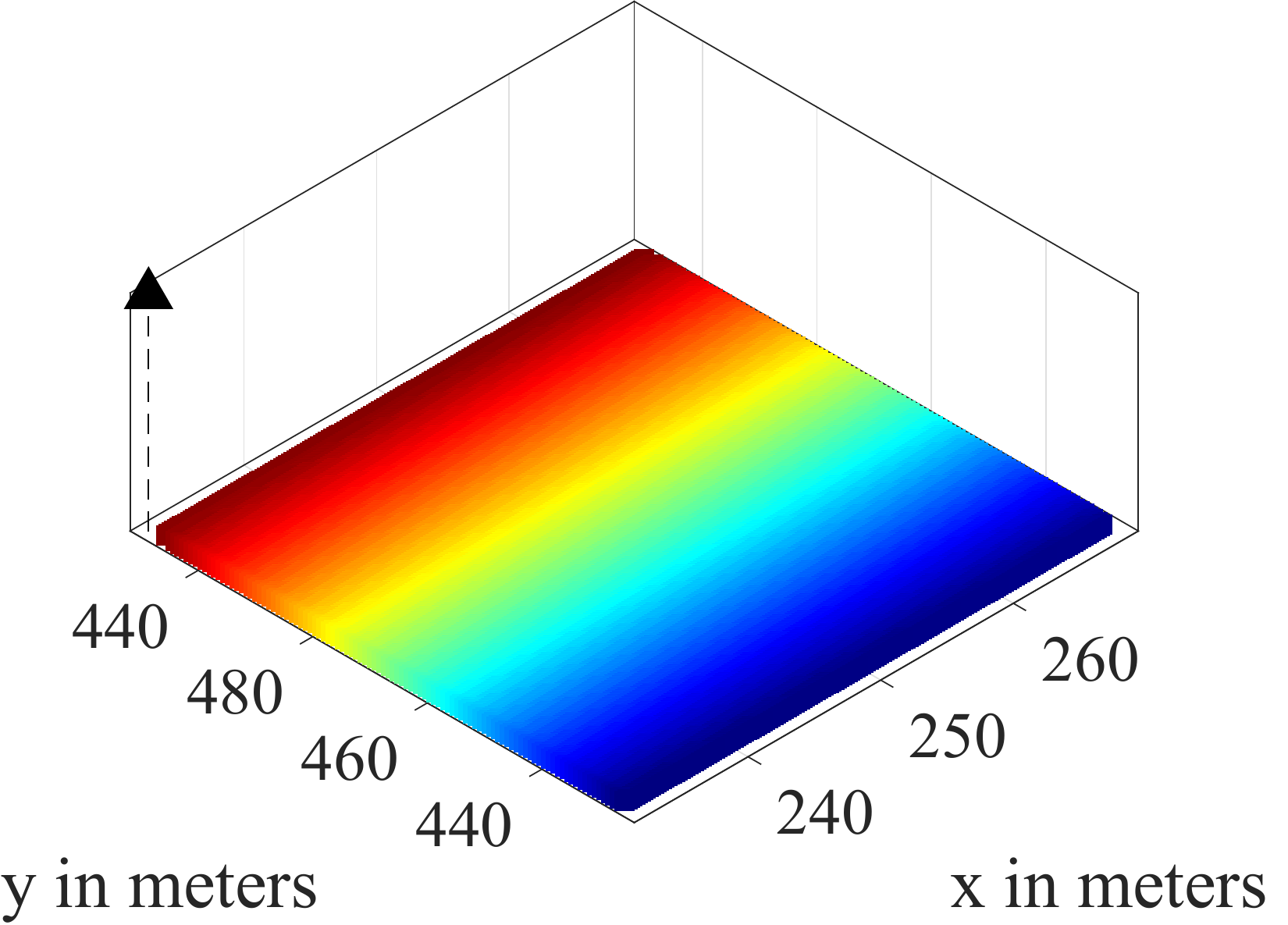}\includegraphics[width=.21\textwidth]{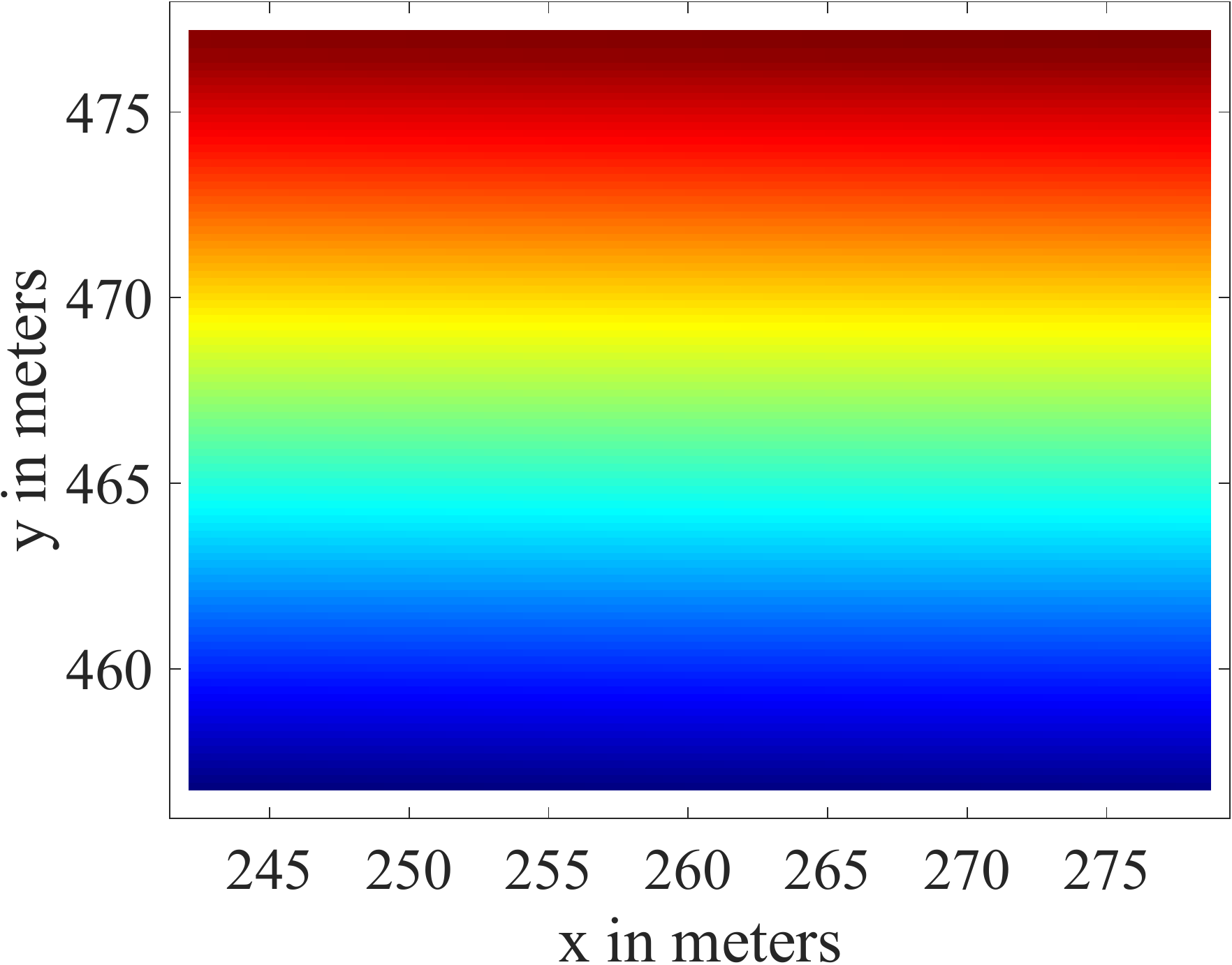}}
\caption{Illustration of the considered DeepMIMO scenario.}
\label{Fig:SA_Illustrate}
\end{figure}

\subsection{Dataset Details and Parameter Settings}
We consider a DeepMIMO scenario where the BS is equipped with ${N} = 32$ antennas operating at 60 GHz with a system bandwidth $W$ of 500 MHz. The UEs are placed on a uniform x-y grid (100 rows and 181 columns, a total of~18,100 distinct spatial locations) with an adjacent spacing of~20 cm between every two UEs. This is illustrated in Fig.~\ref{Fig:SA_Illustrate} (3D and 2D view), where for easy visualization, we use different colors to represent distinct spatial locations in the SA. The black triangle corresponds to the BS location. We stress that the choice of a simple scenario and a well-structured dataset is biased in favor of showing the proof of concept\footnote{Python script to generate the results along with the considered RT data are available online at \url{https://research.ece.ncsu.edu/mpact/data-management/}}. {In~the DeepMIMO dataset, the considered scenario corresponds to a specification of~{BS~3} and {R800} to~{R900}~\cite{Alkhateeb2019}. For the feature transformation (virtual representation), we sample the AoA and delay domain with $N_\text{D} = 32$ and $T_\text{D} = 60$, respectively. Finally, we generated the channel samples (element-space and virtual-space) synthetically based on the discussed parameter values.}\looseness = -1  
 
For the UMAP settings, we set the output space to be~2D ($dim = 2$ for visualization purpose) and the distance metric~$d$ to be Euclidean norm. Choosing the other two hyper-parameters $n$ and $d_{\min}$ depends on the goal (projection of the embedding), and thus, we vary these as free-parameters and run UMAP multiple times. Up next, we first evaluate the efficacy of UMAP with no CHs to see how the embedding projects for various hyper-parameters.\looseness = -1

\subsection{UMAP with no Coverage Hole}
\begin{figure}[t]
\begin{tabular}{c}
\hspace{-.4cm}
{\centering
\includegraphics[width=.13\textwidth]{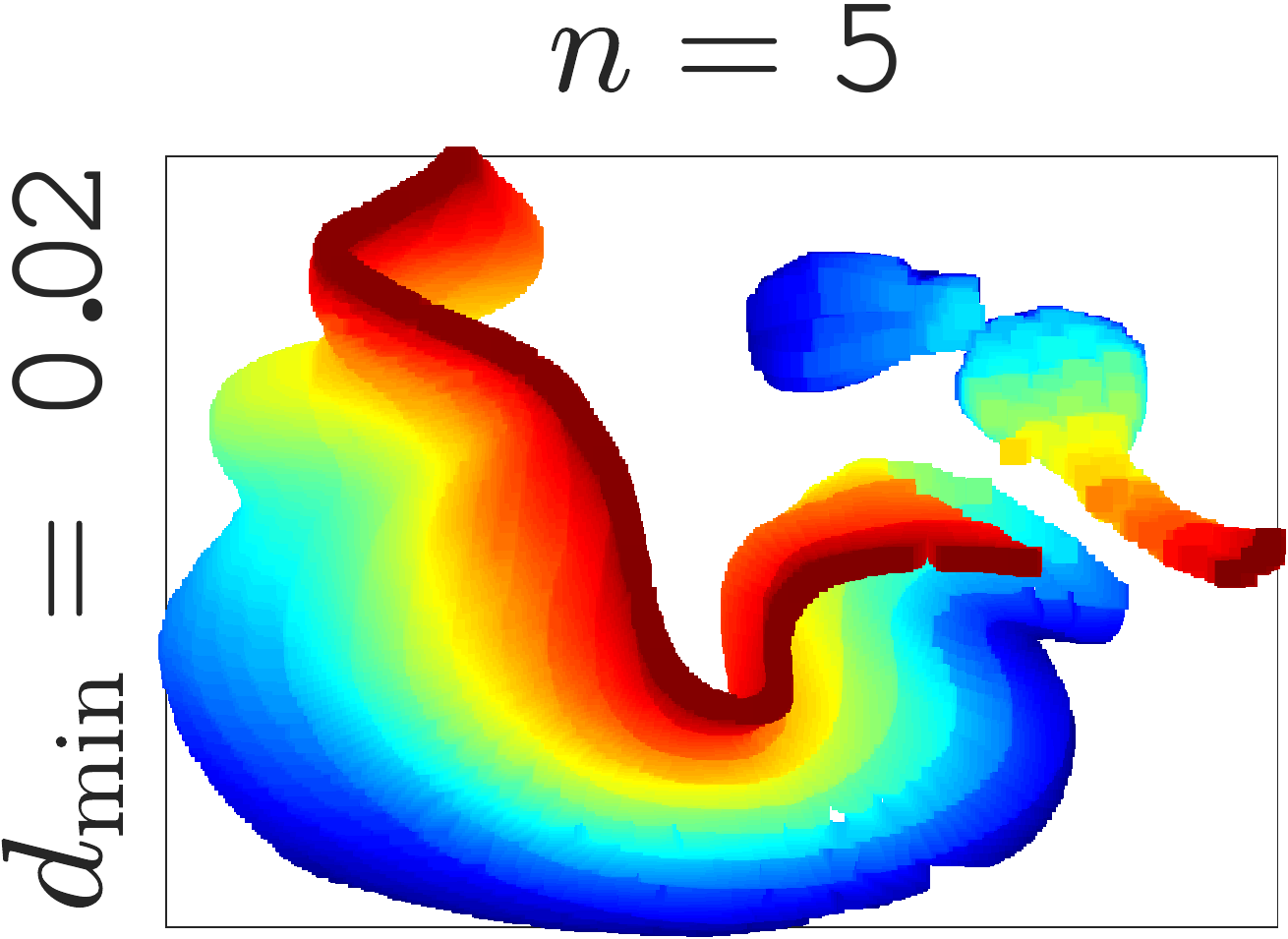}
\includegraphics[width=.115\textwidth]{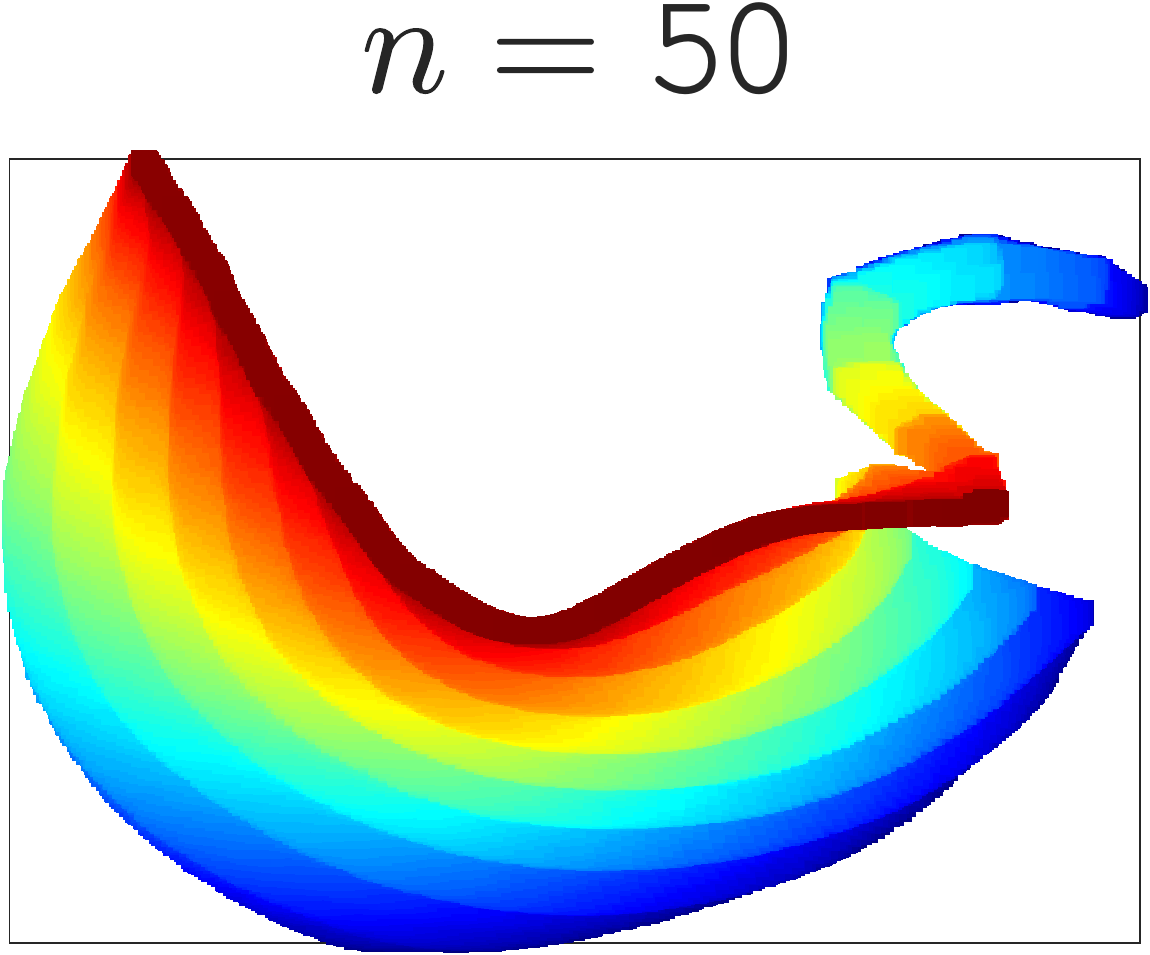}
\includegraphics[width=.115\textwidth]{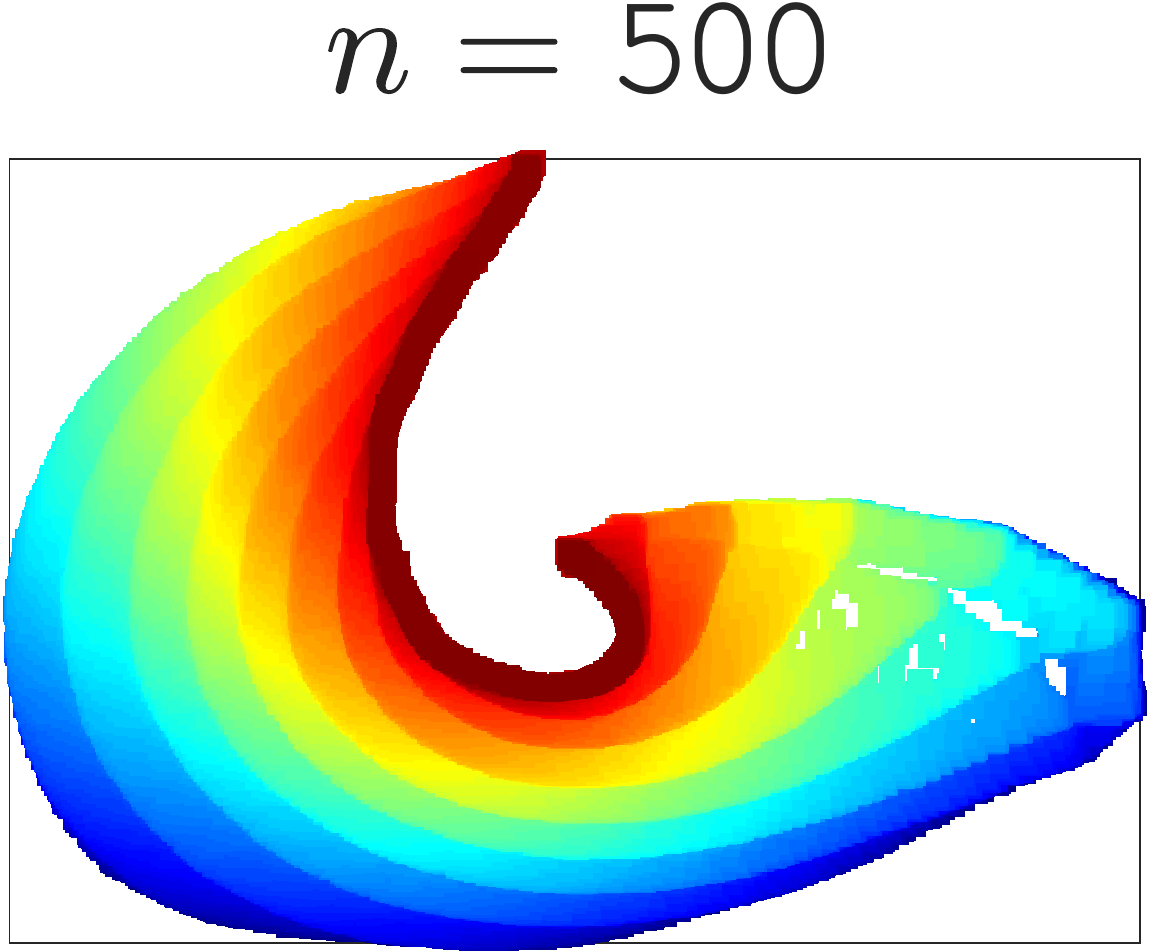}
\includegraphics[width=.115\textwidth]{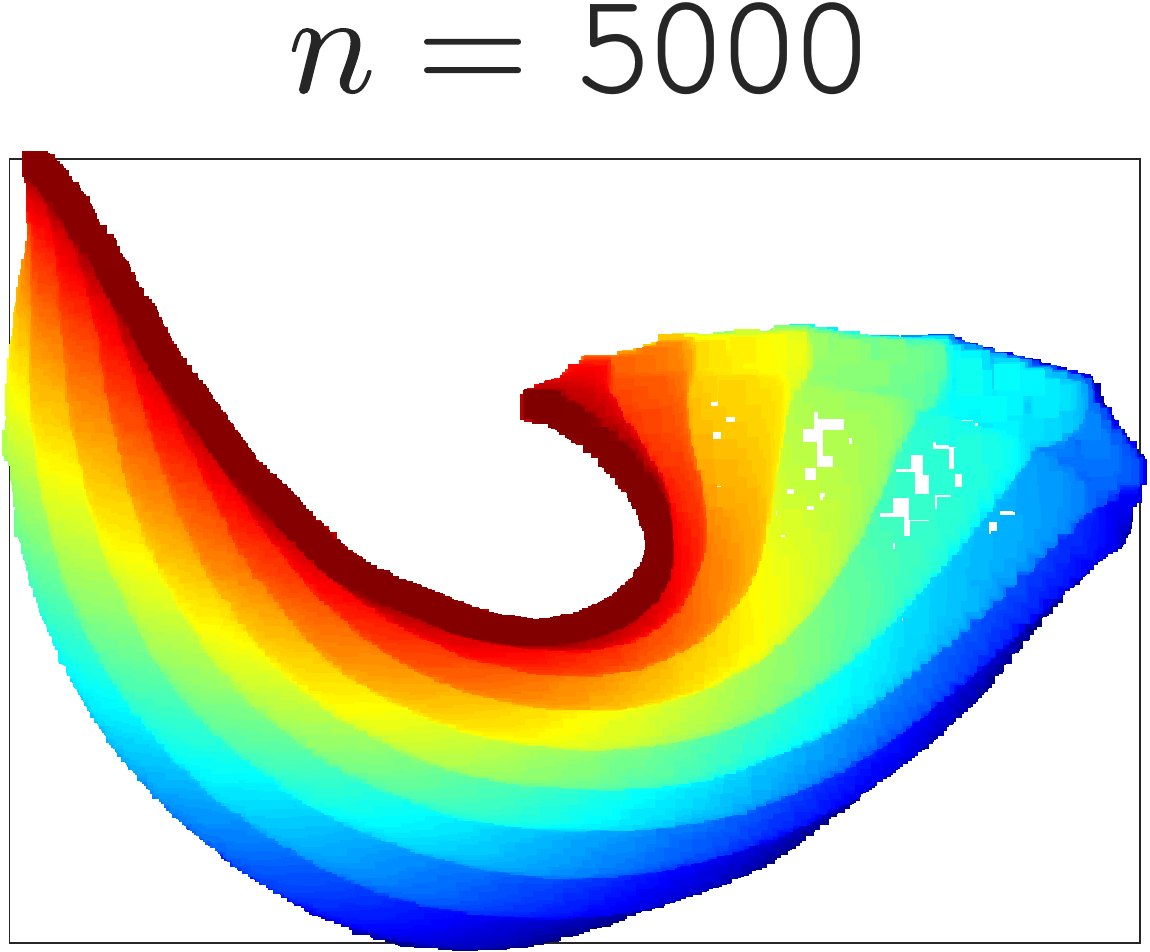}}
\\
\hspace{-.4cm}
{\centering
\includegraphics[width=.13\textwidth]{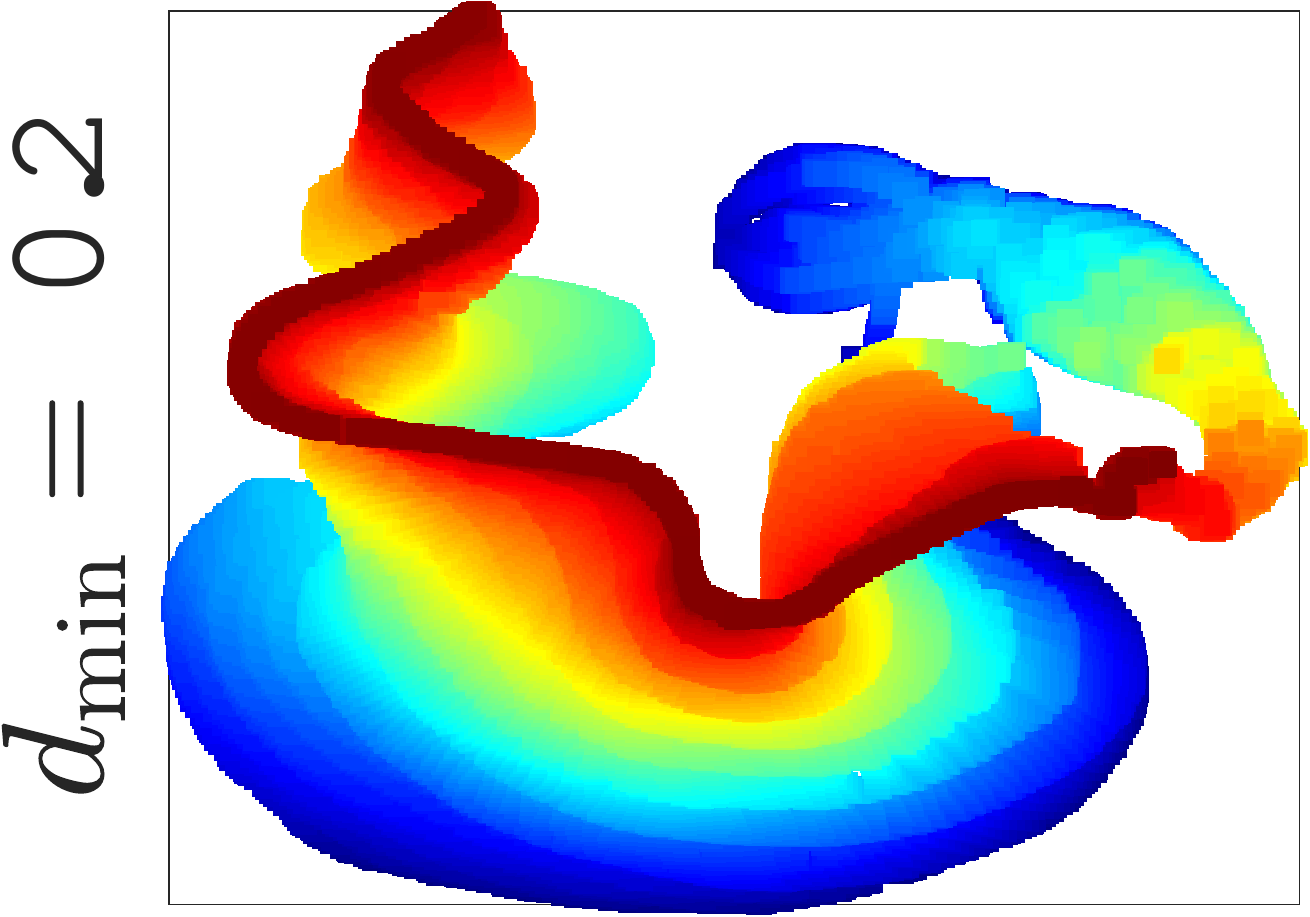}
\includegraphics[width=.115\textwidth]{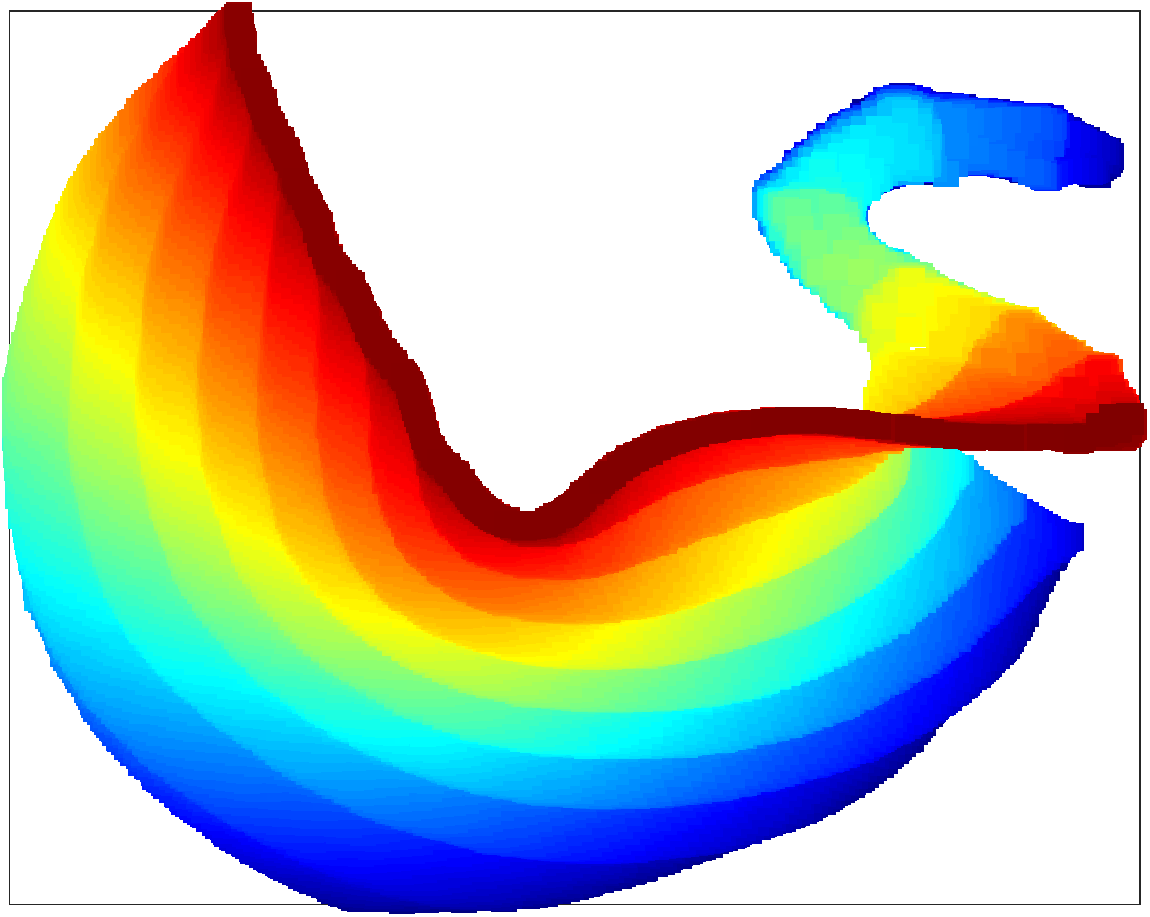}
\includegraphics[width=.115\textwidth]{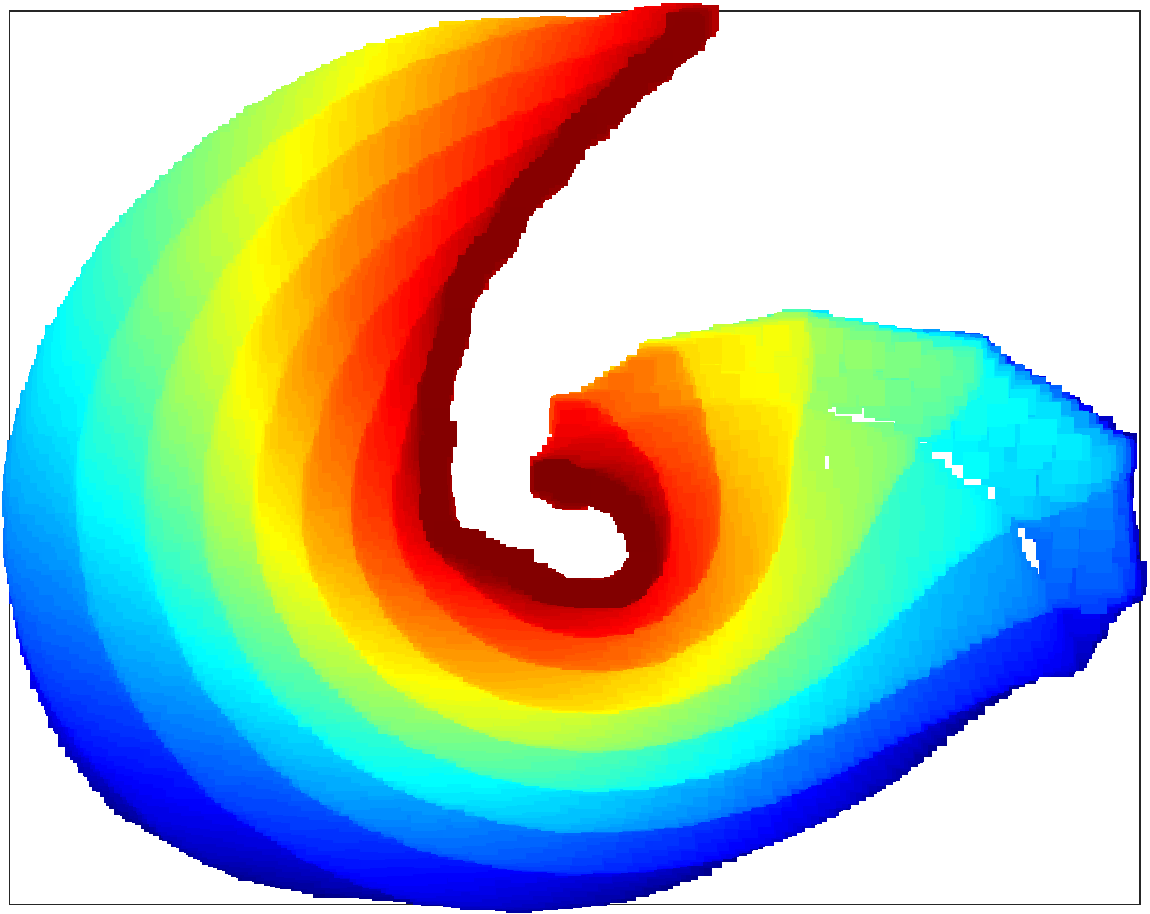}
\includegraphics[width=.115\textwidth]{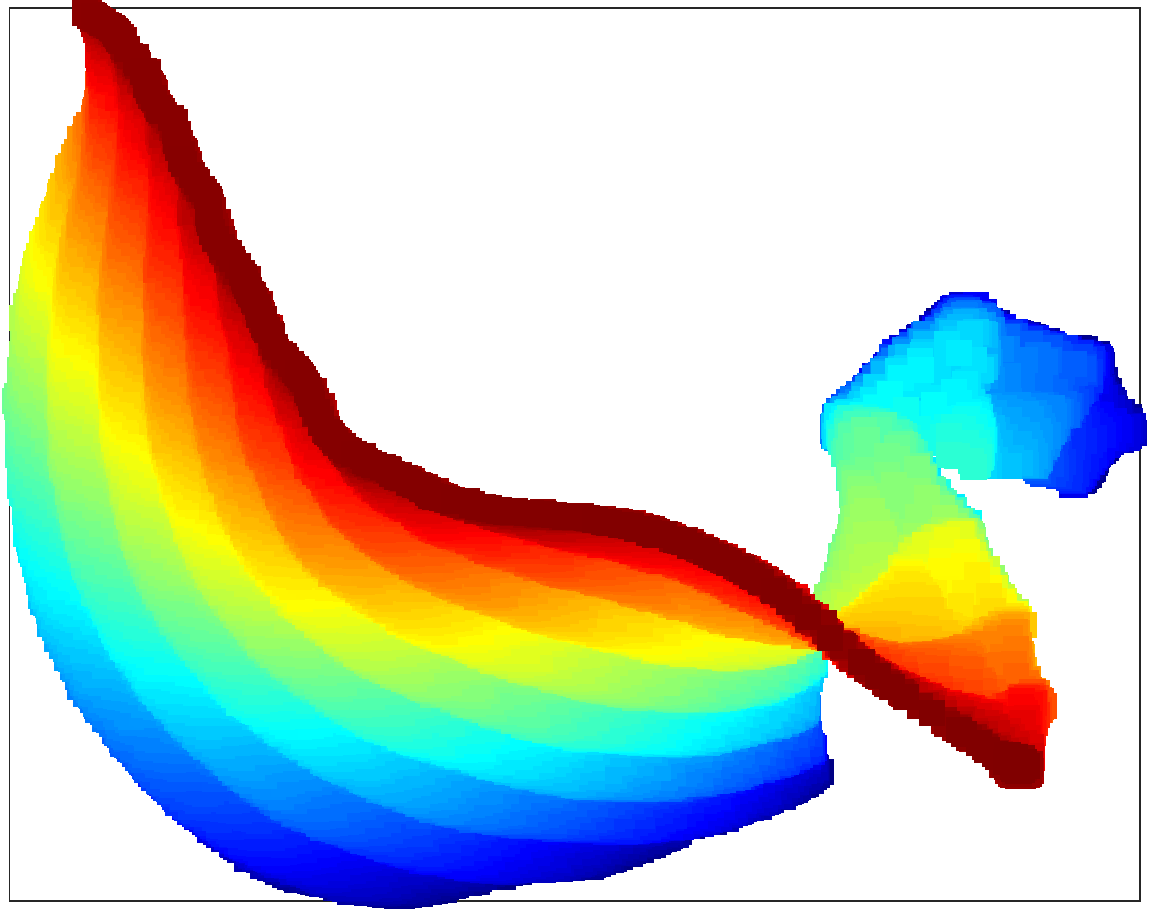}}
\\
\hspace{-.4cm}
{\centering
\includegraphics[width=.13\textwidth]{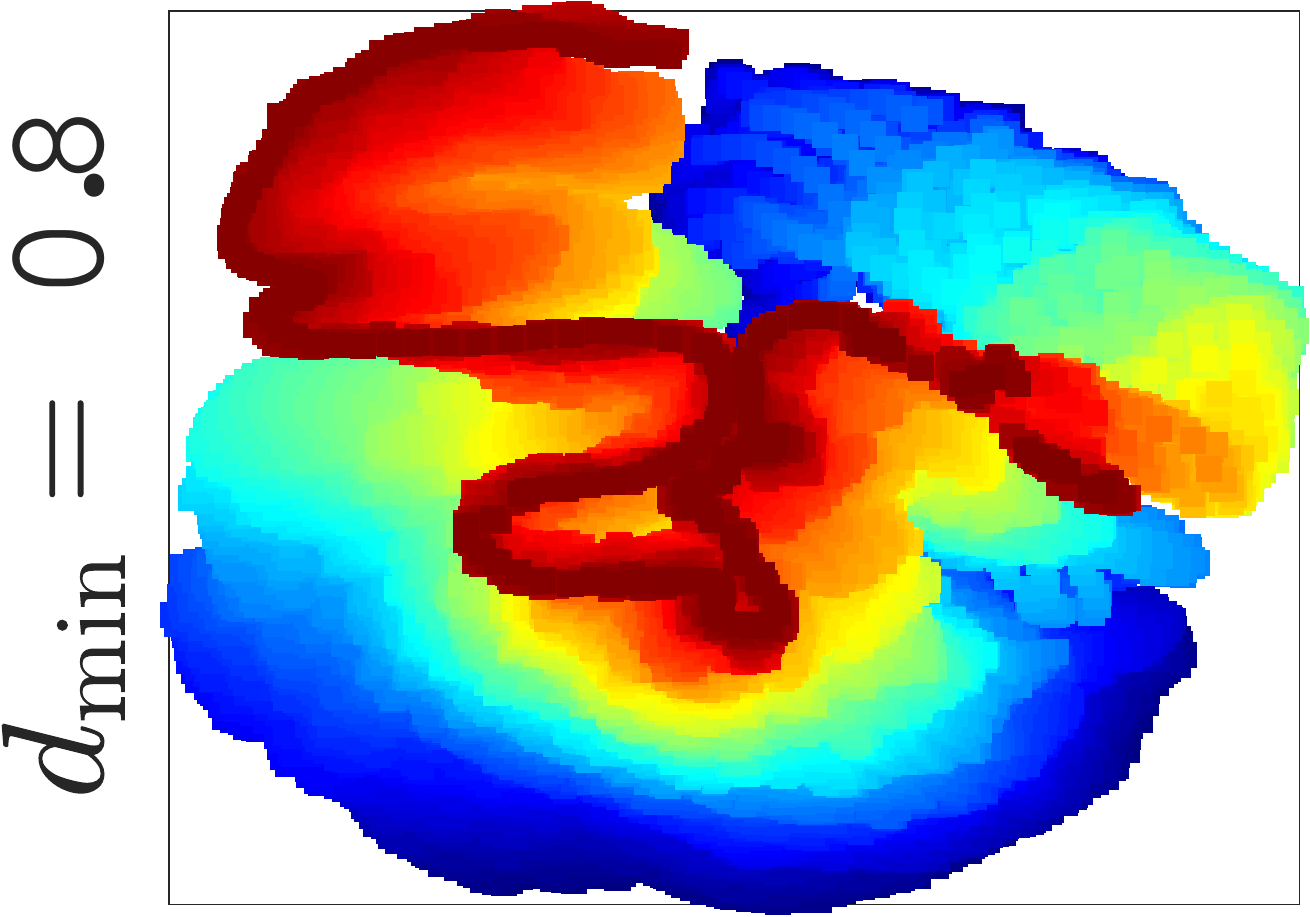}
\includegraphics[width=.115\textwidth]{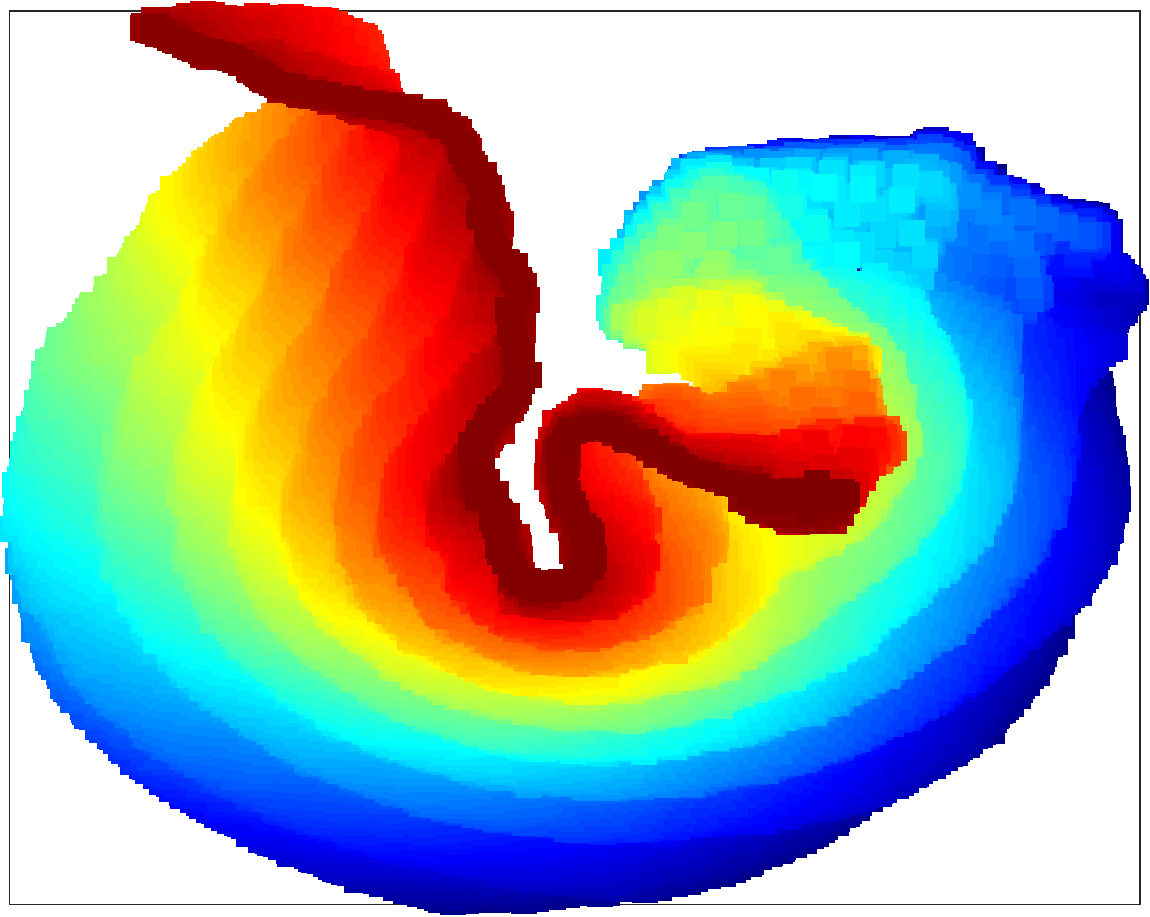}
\includegraphics[width=.115\textwidth]{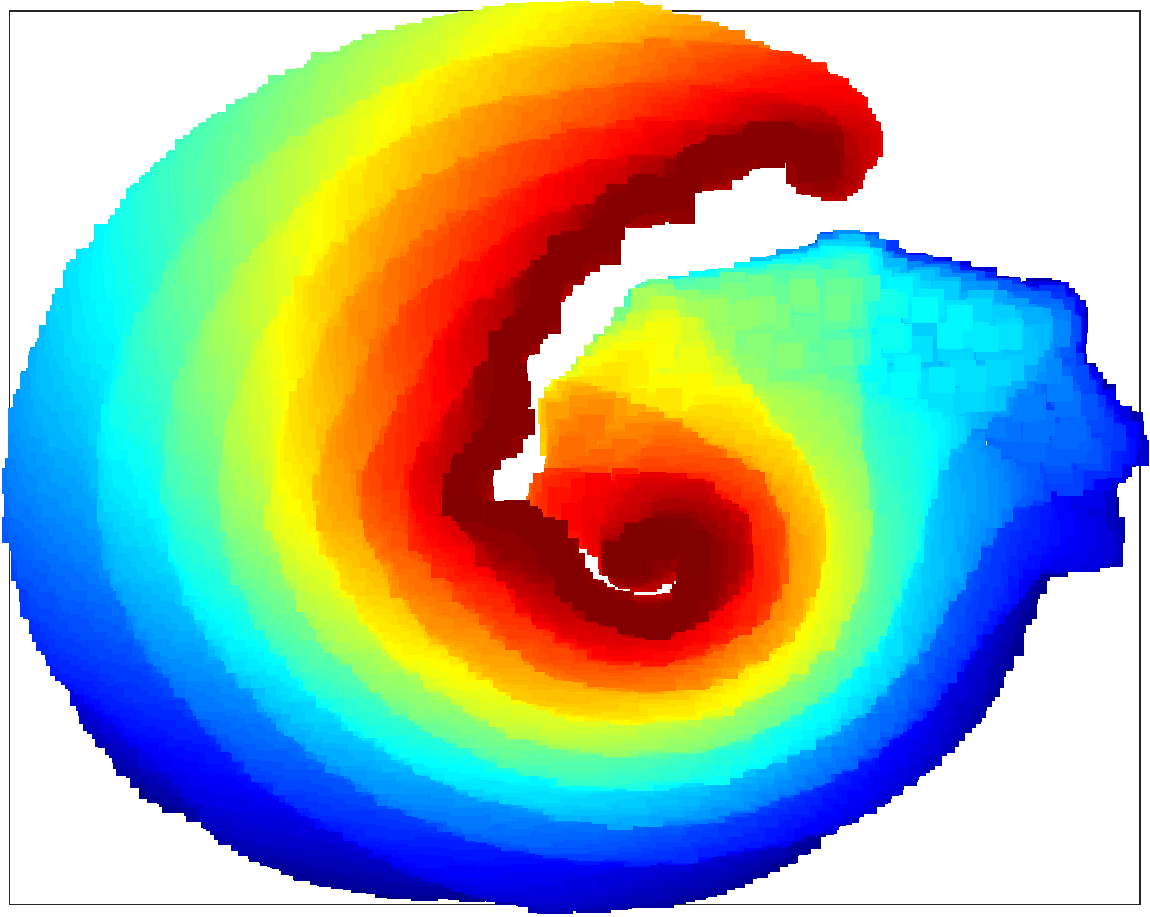}
\includegraphics[width=.115\textwidth]{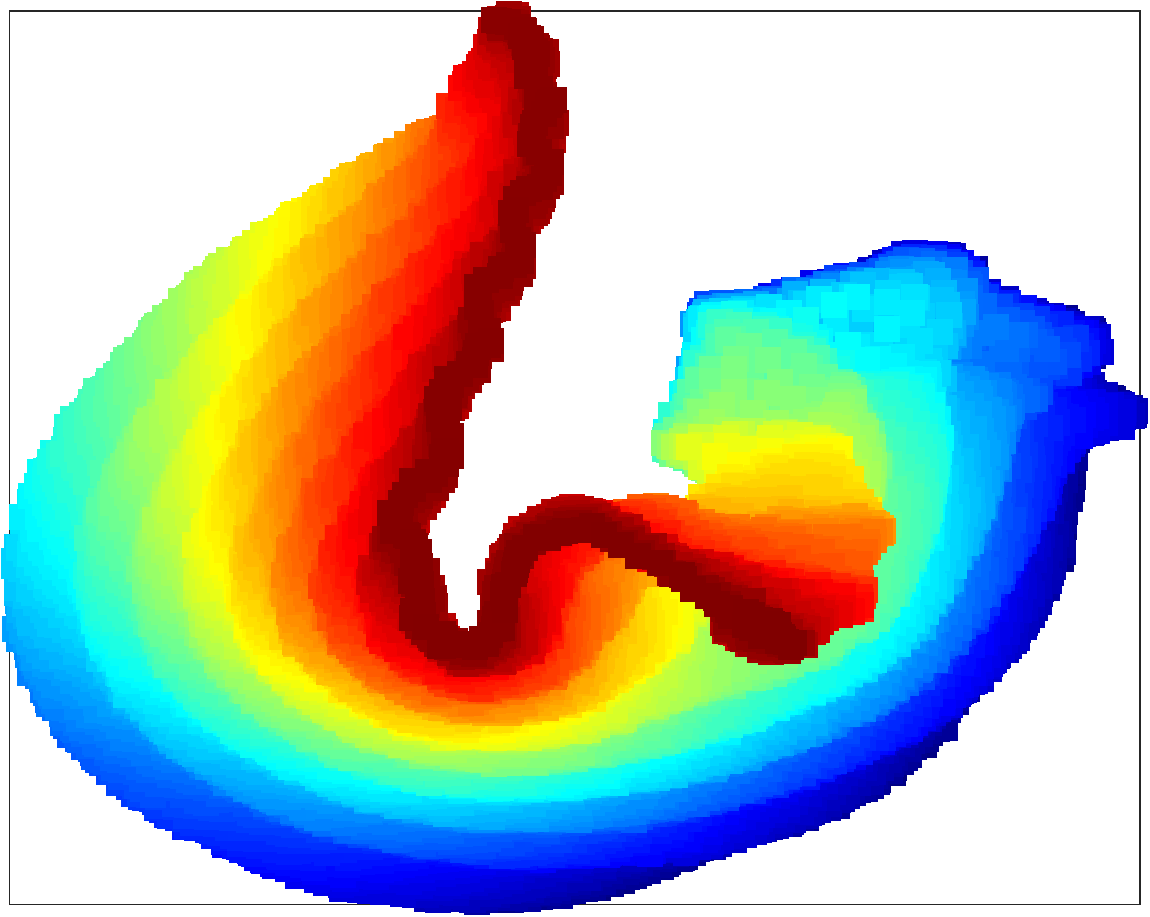}}
\end{tabular}
\caption{2D UMAP embedding for different values of $n$ and $d_{\min}$.}
\label{fig:UMAP_no_CH}
\end{figure}

Fig.~\ref{fig:UMAP_no_CH} illustrates the 2D UMAP embedding for a different choice of $n$ and $d_{\min}$ for the SA (with no CH) shown in~Fig.~\ref{Fig:SA_Illustrate}. We deliberately ignore the low-dimensional embedding axis values as it has no physical meaning. As evident from Fig.~\ref{fig:UMAP_no_CH}, for the right balance of the $n$ and $d_{\min}$, UMAP preserves the structure of the SA in a visually nice fashion (see result for $n=500$ and $d_{\min} = 0.02$ and $0.2$). At a low-value of $n$, the behavior of local-connectedness is prevalent, and increasing $n$ leads to capturing more neighbors leading to better preservation of the global structure. Similarly, high value of $d_{\min}$ leads to compact representation and a converse effect for low values of $d_{\min}$.\looseness = -1

{The curling nature of the projection is evident at a higher value of $d_{\min}$ and lower value of $n$ which is due to the considered SA structure, where the red points are too close to the BS. Enforcing lower~$n$, combined with high $d_{\min}$, would enforce points to be curled to maintain the neighborhood structure compactly. Also, not surprisingly, there might be visual holes even in a continuous space (see the result for $n=500$ and $d_{\min} = 0.02$), but this will vanish with the suitable selection of hyper-parameters} indicating they are not possible CHs. For the considered setup, we noticed medium-to-high values of $n$ and the medium values of $d_{\min}$ effectively capture-and-preserve inherent structure in the data. From the above perspective, it becomes clear why tuning the hyper-parameters is necessary.\looseness = -1

\subsection{UMAP with Coverage Hole(s)}
\begin{figure}[!t]
   \centering
   \subfloat[SA with single CH.]{\includegraphics[width=.1625\textwidth]{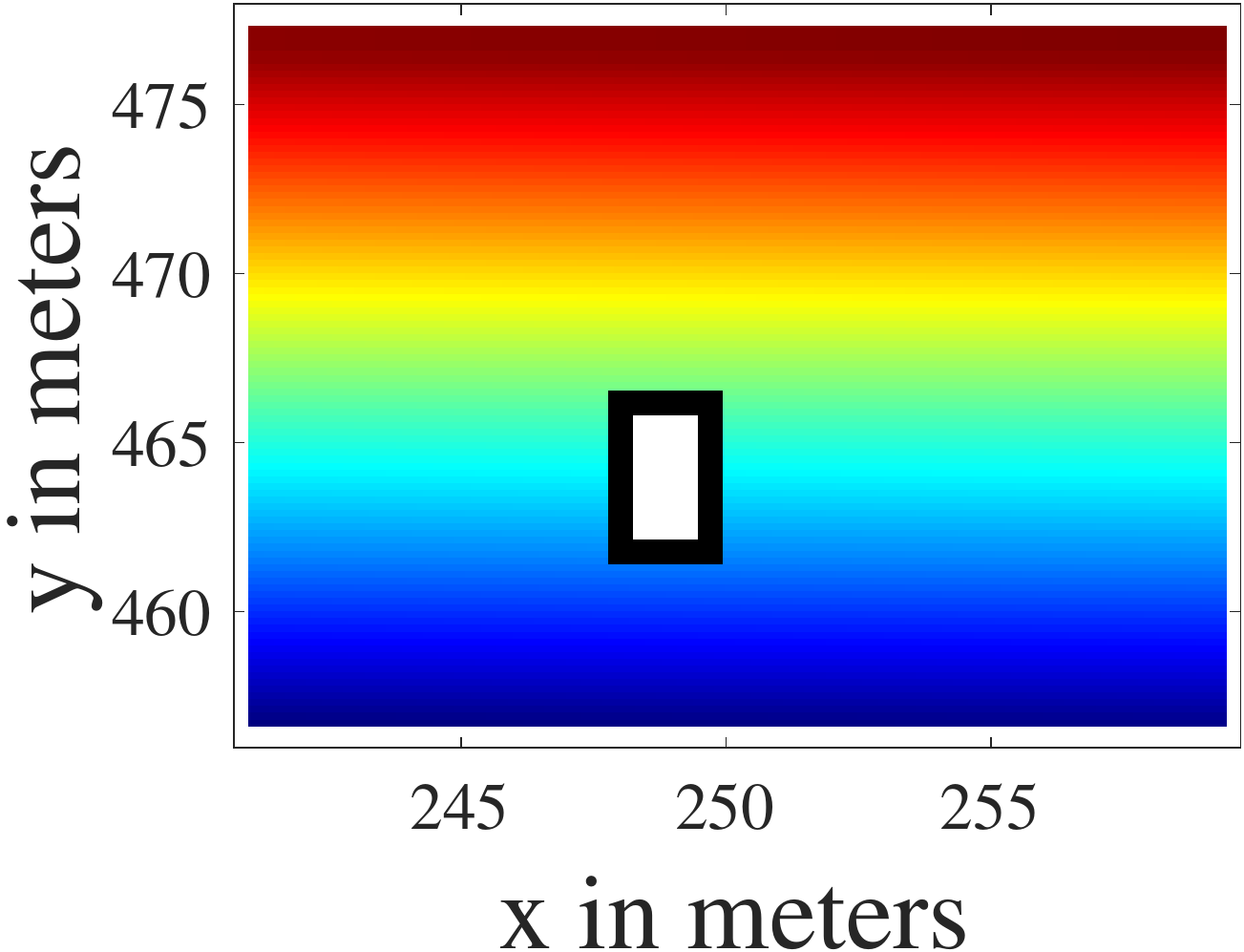}}
   \subfloat[\scriptsize PCA: $\mathcal{T}{\min}$ = 0.32.]{\includegraphics[width=.1625\textwidth]{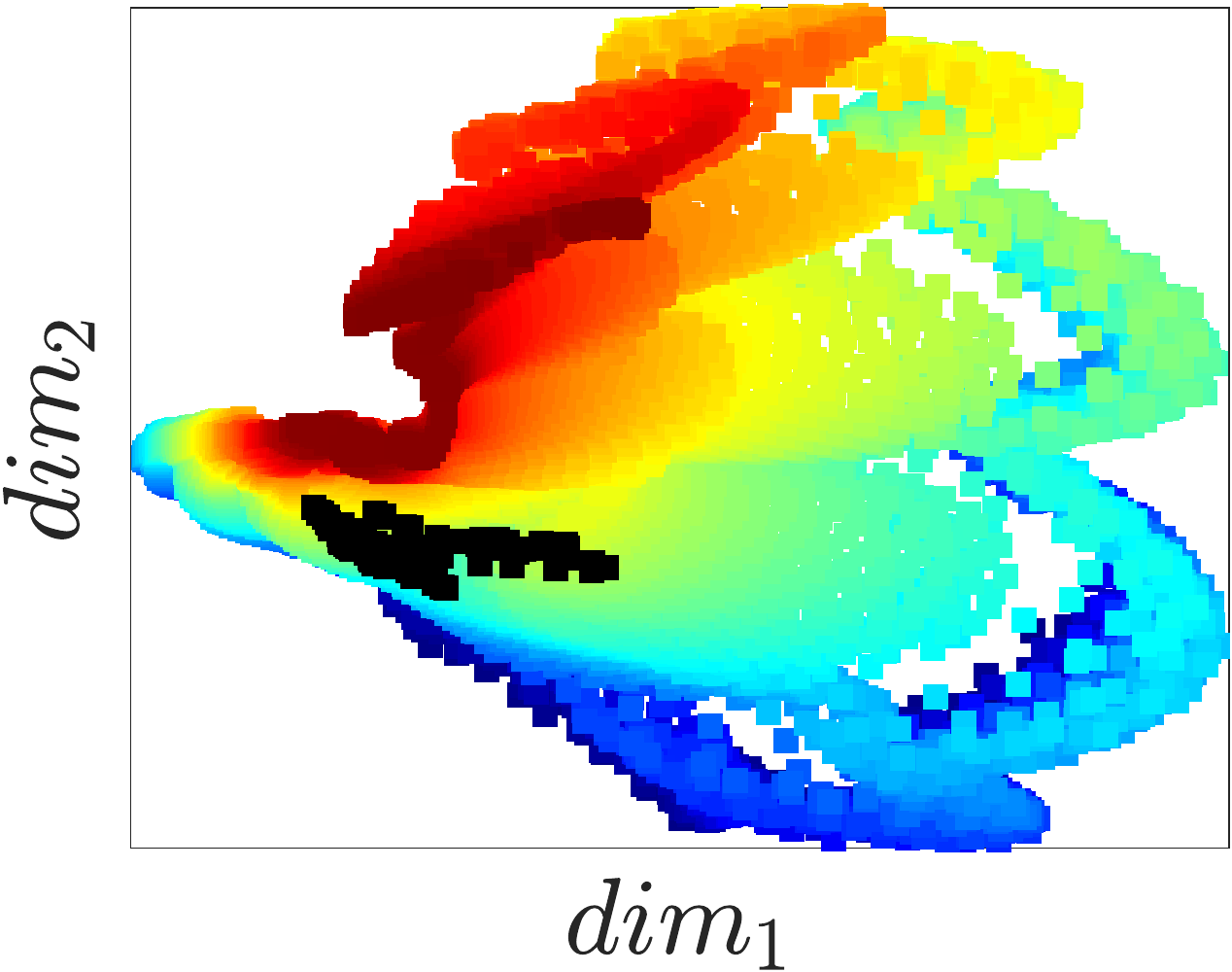}}
   \subfloat[\scriptsize SE: $\mathcal{T}{\min}$ = 0.41.]{\includegraphics[width=.1625\textwidth]{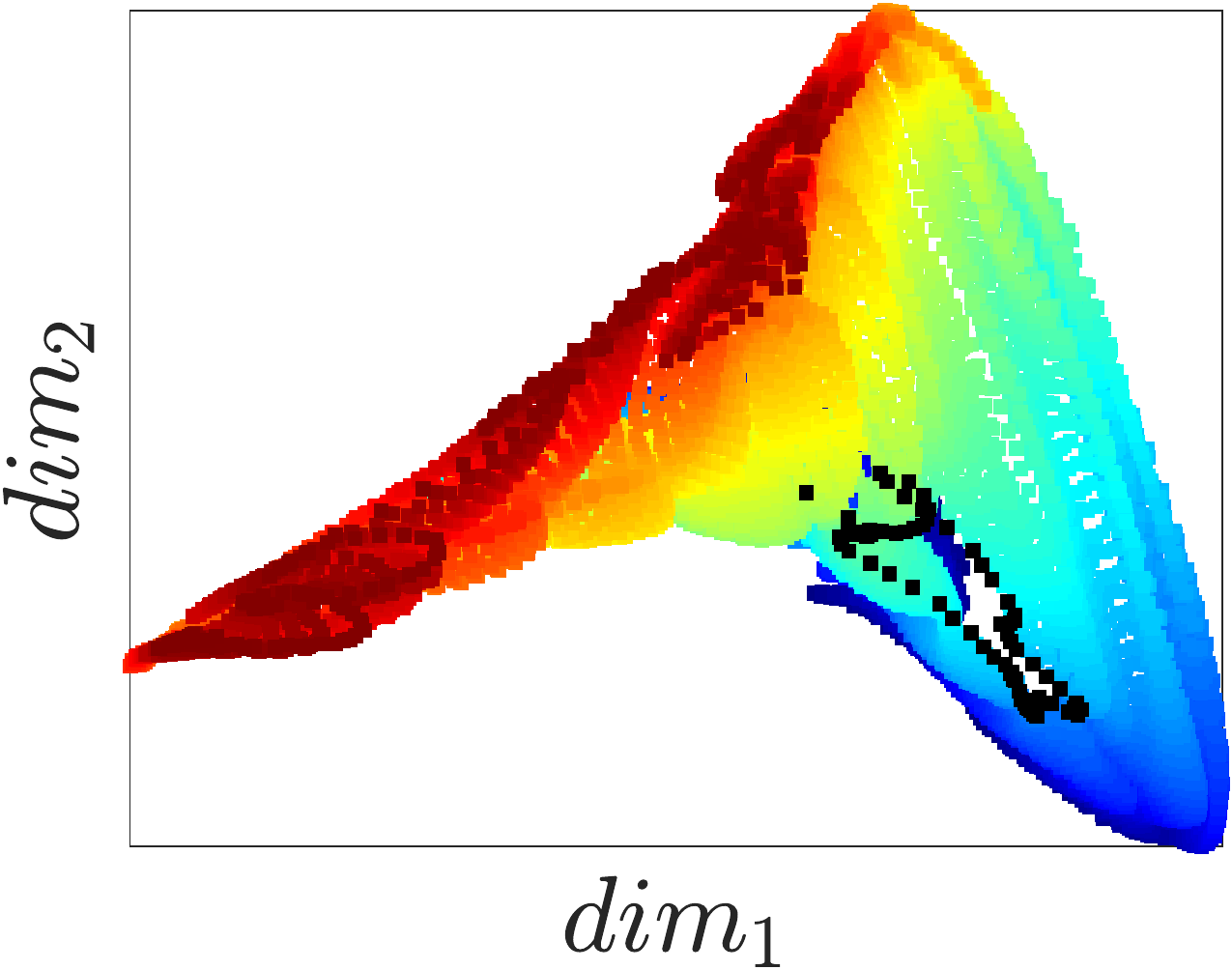}}\\
   \subfloat[\scriptsize LLE: $\mathcal{T}{\min}$ = 0.75.]{\includegraphics[width=.1625\textwidth]{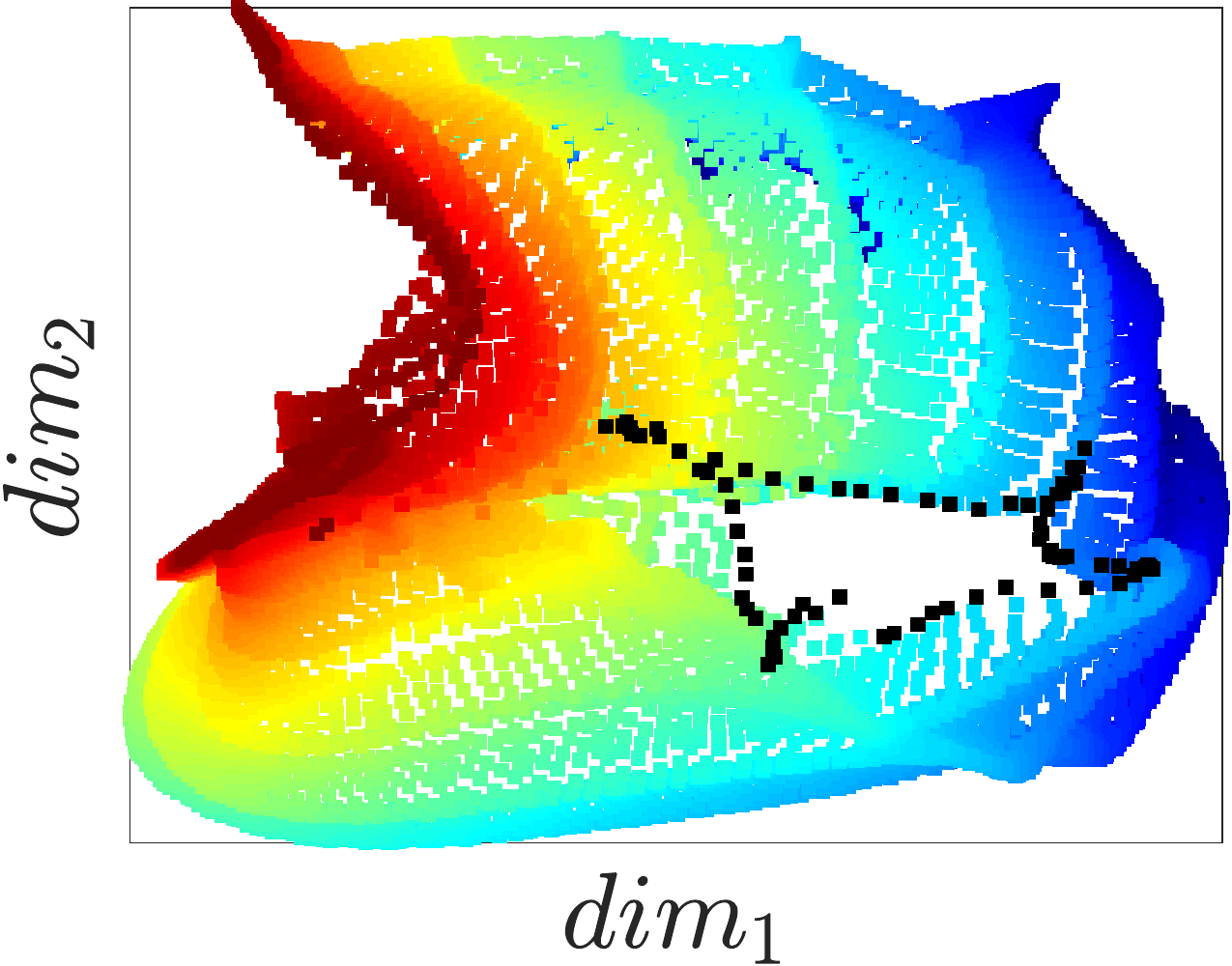}}
    \subfloat[\scriptsize t-SNE:$\mathcal{T}{\min}$ = 0.31.]{\includegraphics[width=.1625\textwidth]{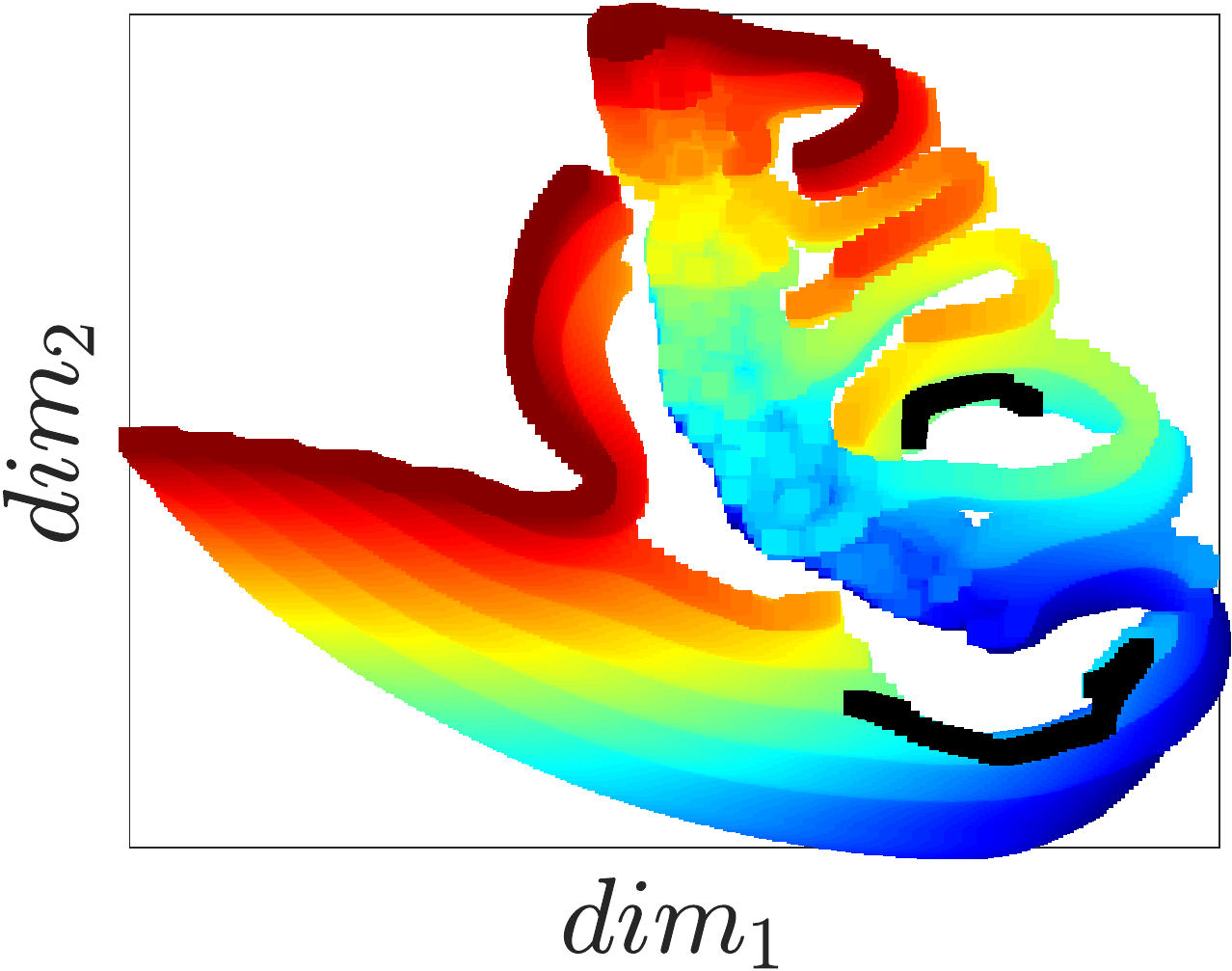}}
    \subfloat[\scriptsize UMAP: $\mathcal{T}{\min}$ = 0.85.]{\includegraphics[width=.1625\textwidth]{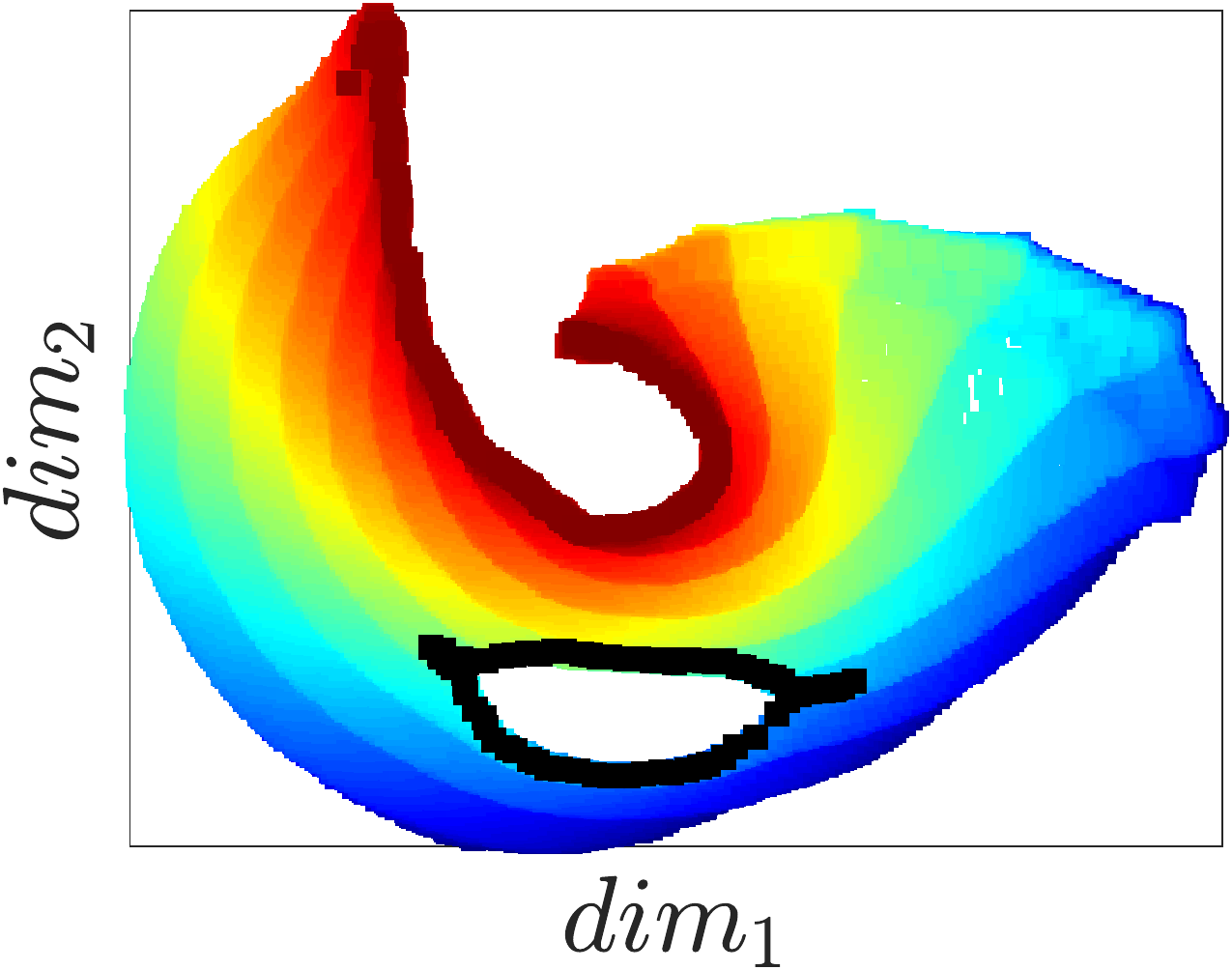}}
   \caption{Single CH: 2D embedding of different algorithms.}
   \label{fig:CH1}
\end{figure}

In this work, we use a synthetic way of simulating CH with a known structure, as simulating actual CHs is difficult. We do so by taking out the channels for a particular region (equivalent to no MPCs in the region), illustrated as the white region in~Fig.~\ref{fig:CH1}(a). {We note that this can be a traffic hole (a region with no traffic); we plan to explore differentiating a CH from a traffic hole in our future work.} In this work, we explicitly assume it as a CH. Further, the CH boundary points (outside the CH) are represented in black color. {For the defined scenario, we present the simulation results from a set of parameters tuned to give good visual embedding. For qualitative comparisons, we compare UMAP ($n =1800, d_{\min} = 0.15$) results against the classical principal component analysis (PCA) and other state-of-the-art dimensional reduction techniques such as t-stochastic neighborhood (t-SNE), spectral embedding (SE), and locally linear embedding (LLE), that act as baselines, all implemented via scikit-learn~\cite{pedregosa2011scikit}. Further, to numerically quantify how well the proximities are preserved, we use the trustworthiness (TW)~\cite{venna2005local} metric (ranges [0,1]) for a fixed neighborhood size of $k =1800$. The point-wise TW term is calculated for each data sample $i\in [M]$, and a minimum of TW denoted by $\mathcal{T}_{\min} = \min_{i} TW(i,k)$ is used as the metric to signify the trustworthiness of the overall result. The optimal value for $\mathcal{T}_{\min} = 1$ signifies no loss of the proximities in the process. } \looseness = -1

{Fig.~\ref{fig:CH1}(b)-(f) demonstrates the 2D embedding of different algorithms along with the $\mathcal{T}_{\min}$ metric for a single CH in the SA illustrated in~\ref{fig:CH1}(a). The UMAP has an apparent visual hole besides capturing the CH boundary smoothly along with the best $\mathcal{T}_{\min}$ values. On the contrary, the PCA and t-SNE fail to capture the CH boundaries, whereas, the SE and LLE do capture the structure to some extent but have lower $\mathcal{T}_{\min}$ values signifying losing some of the proximities in the process. Further, the SE and LLE also have somewhat loose representation, resulting in multiple visual holes that can be wrongly interpreted as CHs. The above phenomenon is because PCA is a linear technique and it does not have the notion of a neighborhood distance capturing mechanism, whereas other considered algorithms do. Although SE, LLE, and t-SNE preserve the neighborhood within the data to an extent, they cluster the data in the visualization as they do not have $d_{\min}$ parameter that leads to super-tight visual representations. On the other hand, the UMAP strikes a balance between local-global structure preservation and provides a super-tight representation in the embedding leading to visual holes.}\looseness = -1

{It is worth noting that the demonstrated t-SNE, LLE, and SE results had similar neighbors as UMAP ({for t-SNE perplexity of $600$}, computes $\approx 1800$ neighbors). With a higher perplexity value, even t-SNE might produce a visual hole in the low-dimensional space. However, this comes with a dramatic increase in run-time. 
UMAP, however, has a shorter run-time than t-SNE for a large dataset and captures the structure even at a lower number of neighbors. Run-times of t-SNE, SE, and LLE increase quadratically with $\mathcal{O}(M^2)$, where $M$ is the number of data-samples. On the other hand, the UMAP's overall complexity is reported empirically, to be approximately $\mathcal{O}(M^{1.14})$~\cite{mcinnes2018umap}. Overall, based on the quality of the results and run-time, it is fair to say that UMAP is a better option.}\looseness = -1

\begin{figure}[!t]
   \centering
   \subfloat[SA with single CH.]{\includegraphics[width=.1625\textwidth]{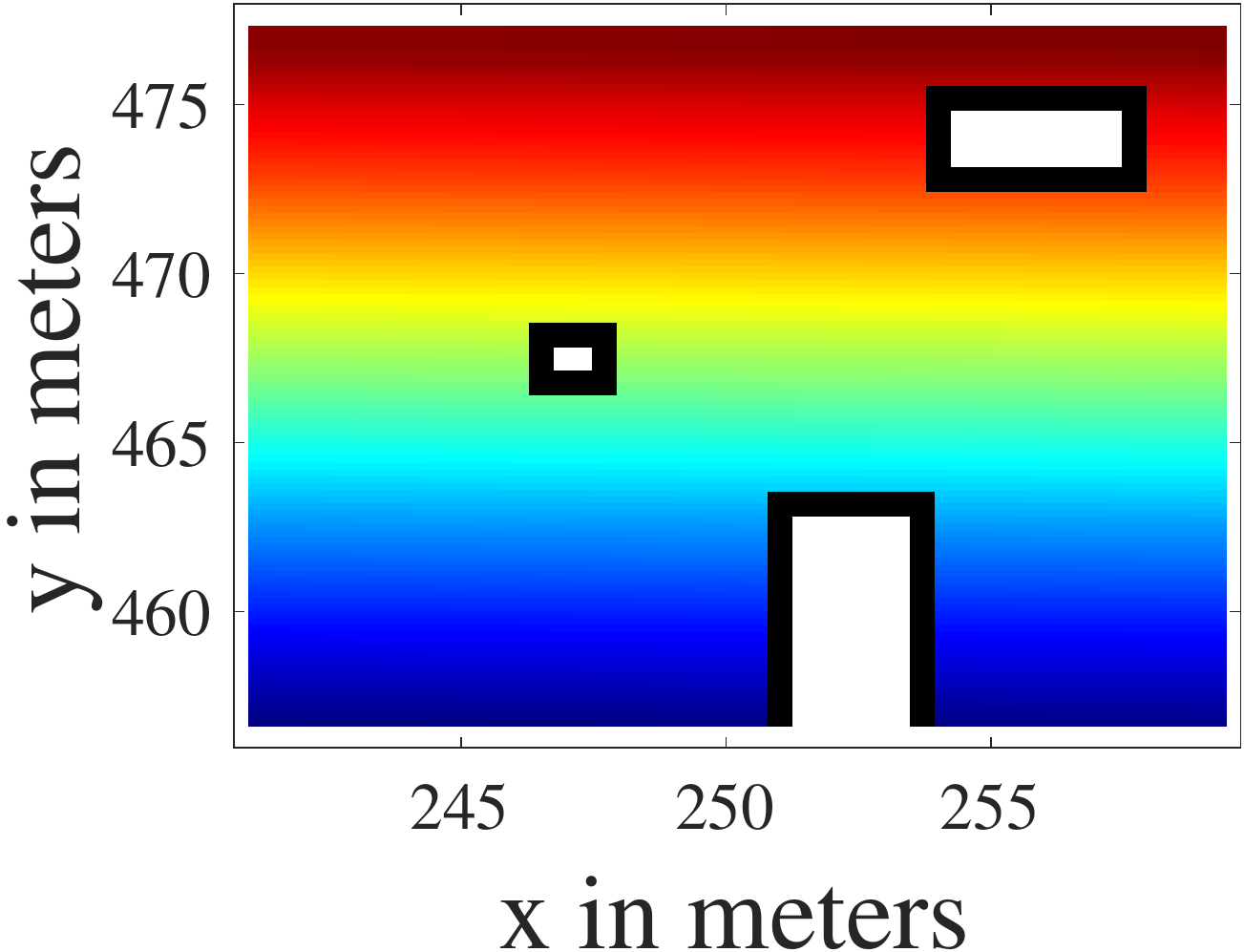}}
   \subfloat[\scriptsize PCA: $\mathcal{T}{\min}$ = 0.39.]{\includegraphics[width=.1625\textwidth]{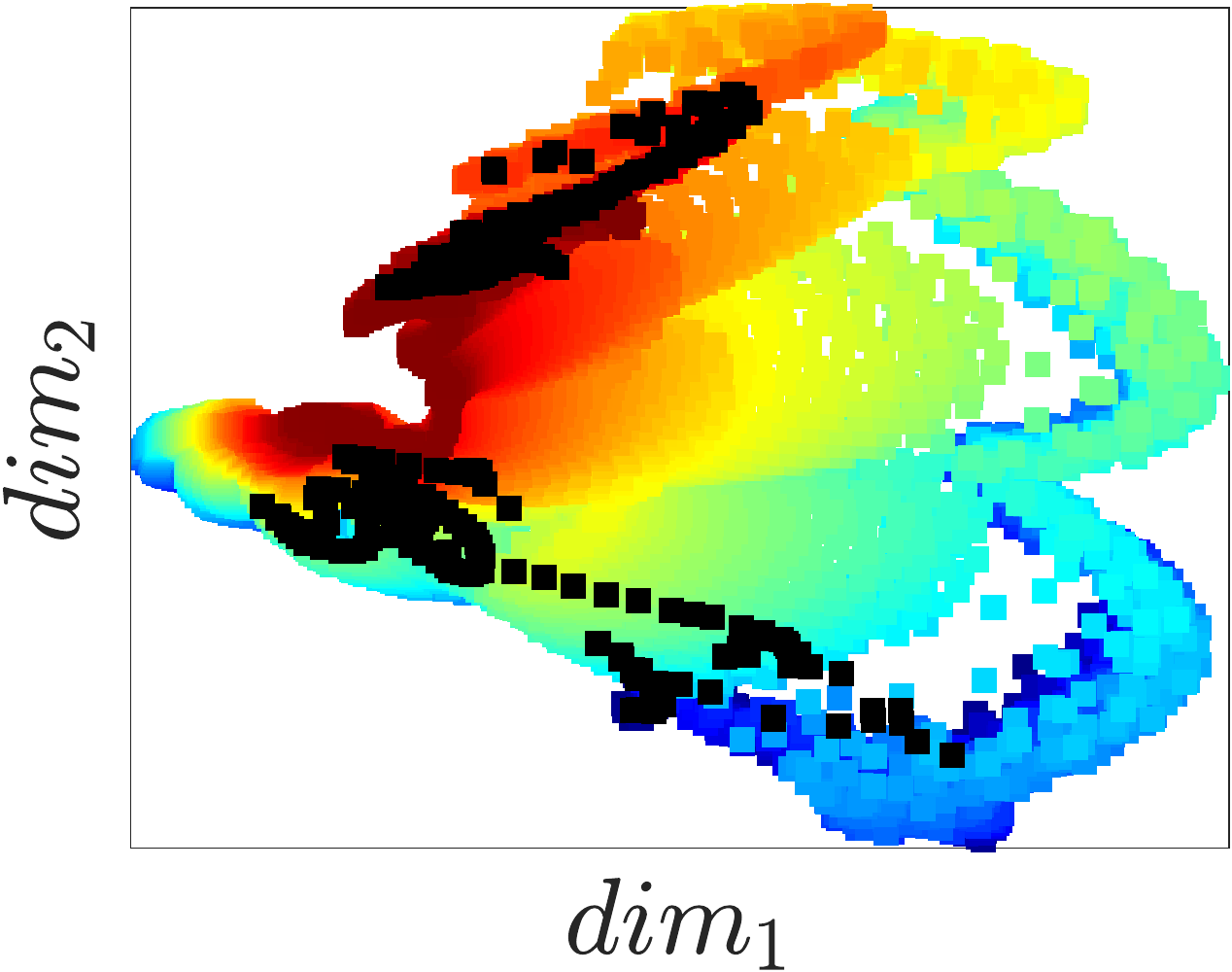}}
   \subfloat[\scriptsize SE: $\mathcal{T}{\min}$ = 0.43.]{\includegraphics[width=.1625\textwidth]{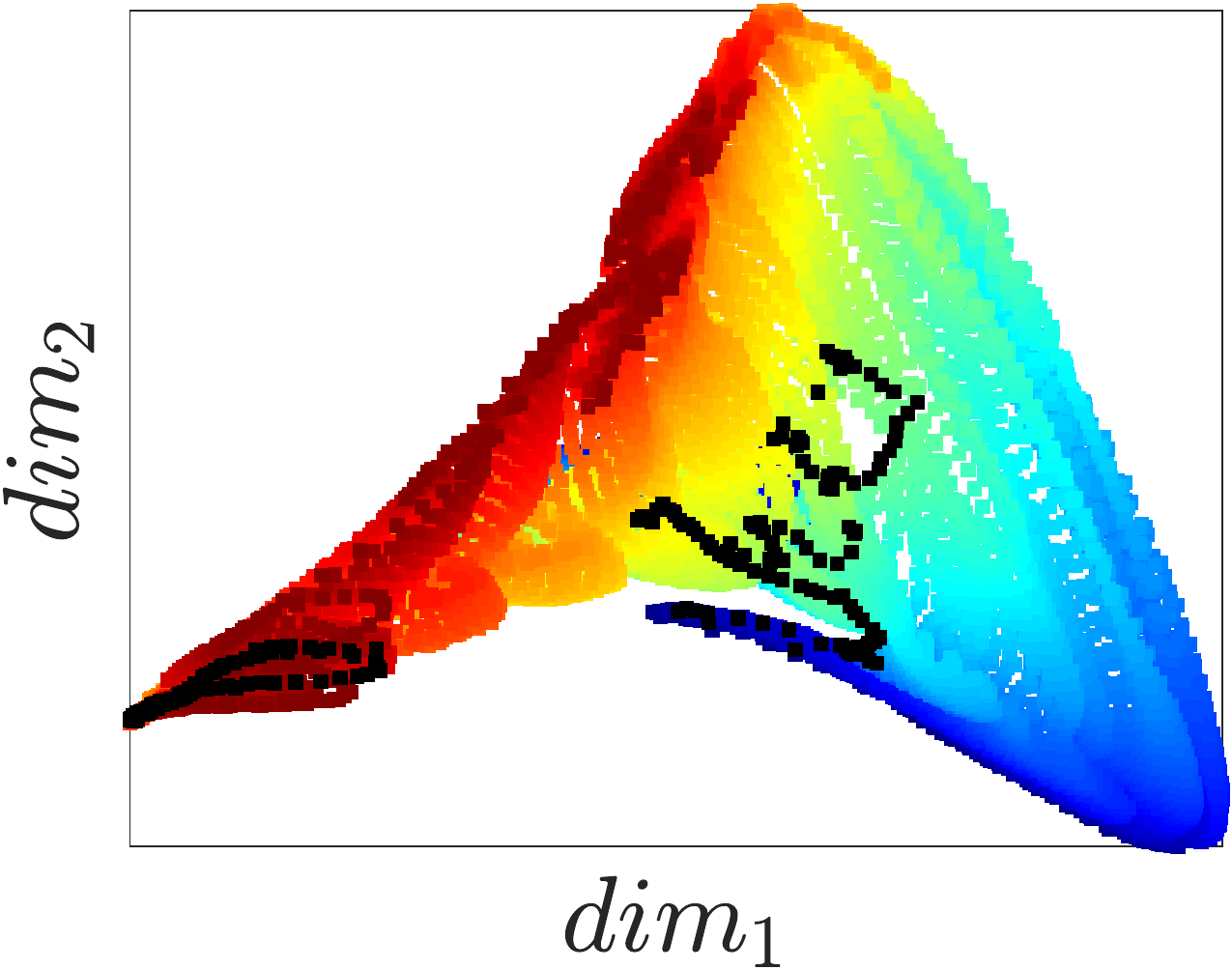}}\\
   \subfloat[\scriptsize LLE: $\mathcal{T}{\min}$ = 0.71.]{\includegraphics[width=.1625\textwidth]{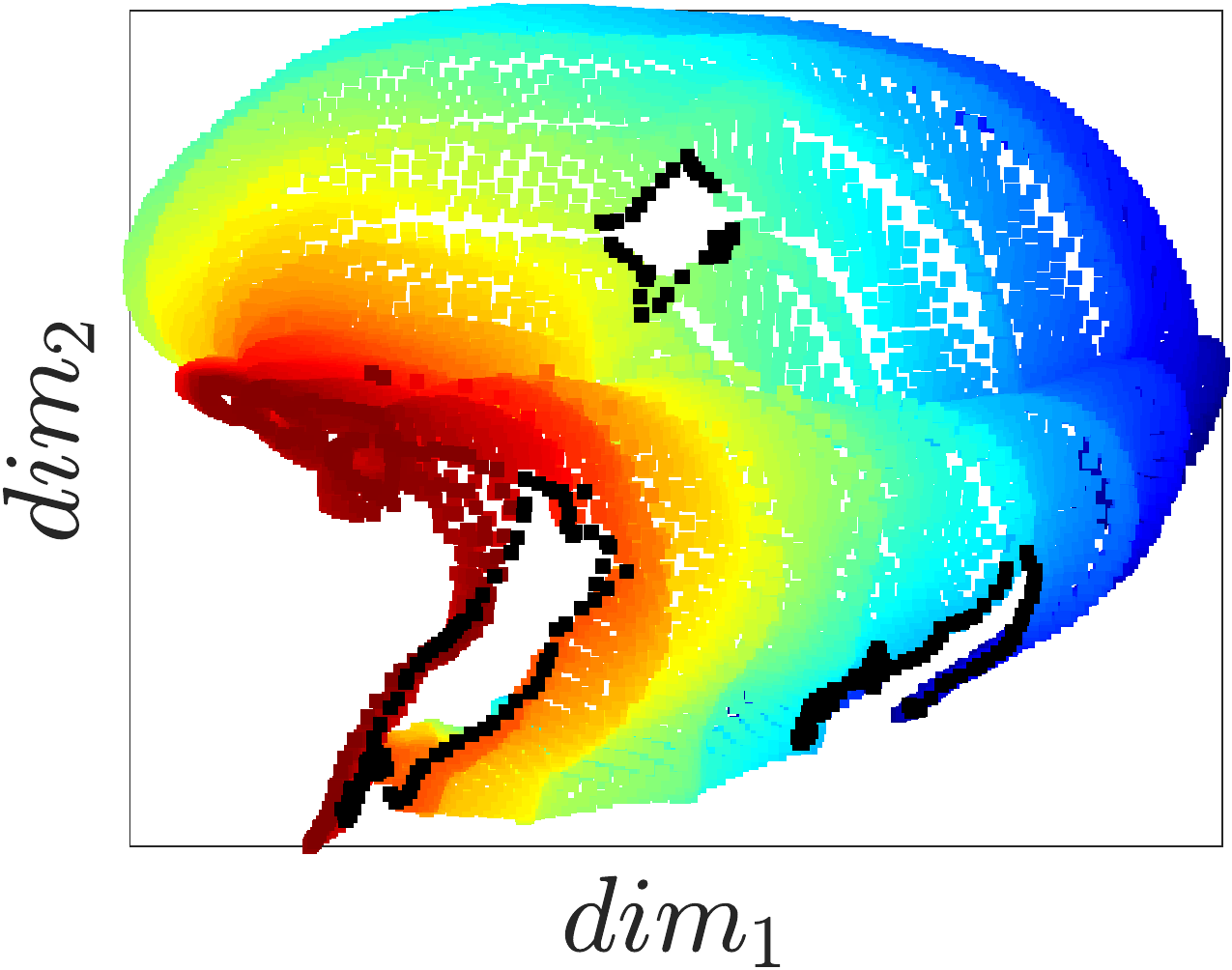}}
    \subfloat[\scriptsize t-SNE: $\mathcal{T}{\min}$ = 0.44.]{\includegraphics[width=.1625\textwidth]{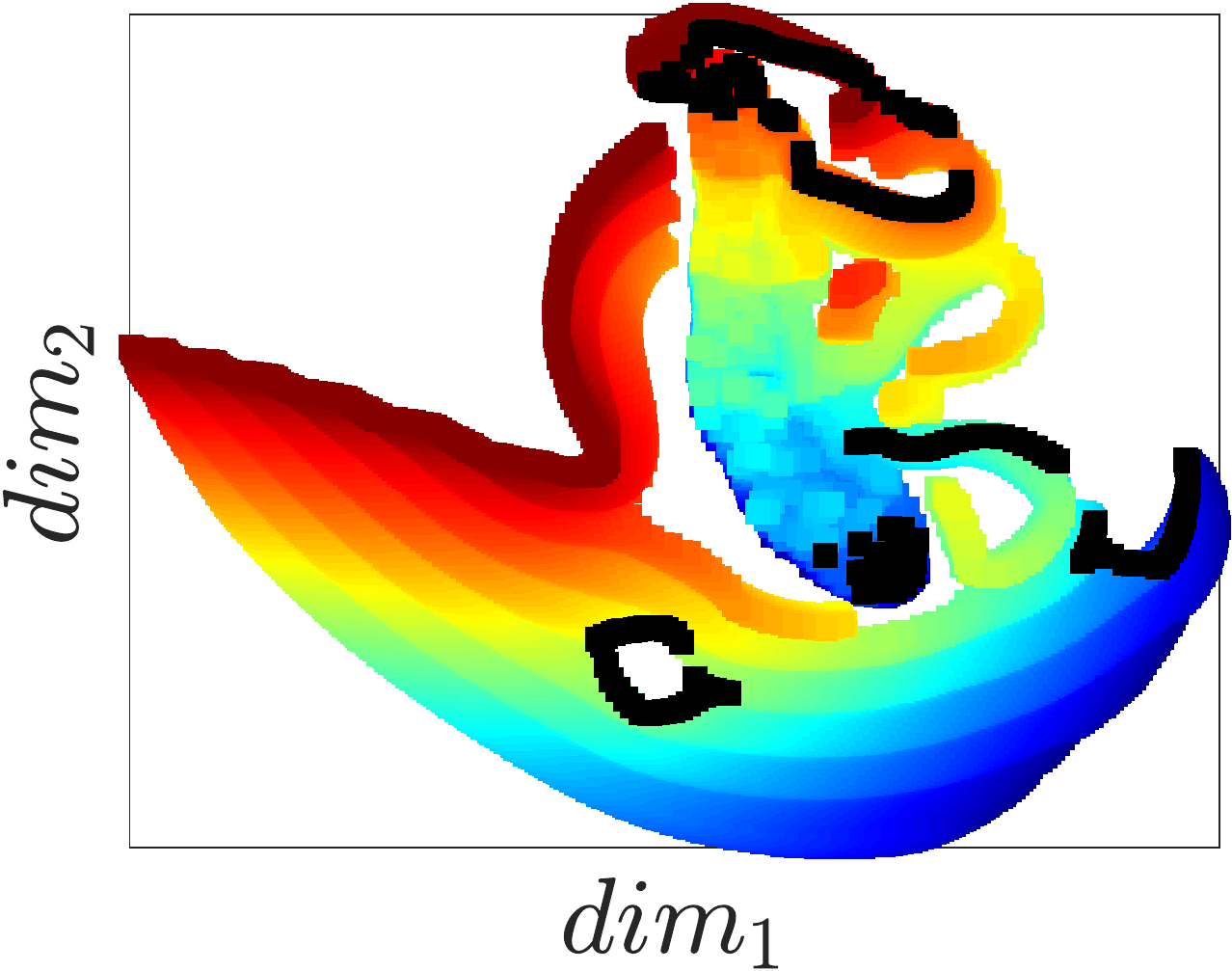}}
    \subfloat[\scriptsize UMAP: $\mathcal{T}{\min}$ = 0.82.]{\includegraphics[width=.1625\textwidth]{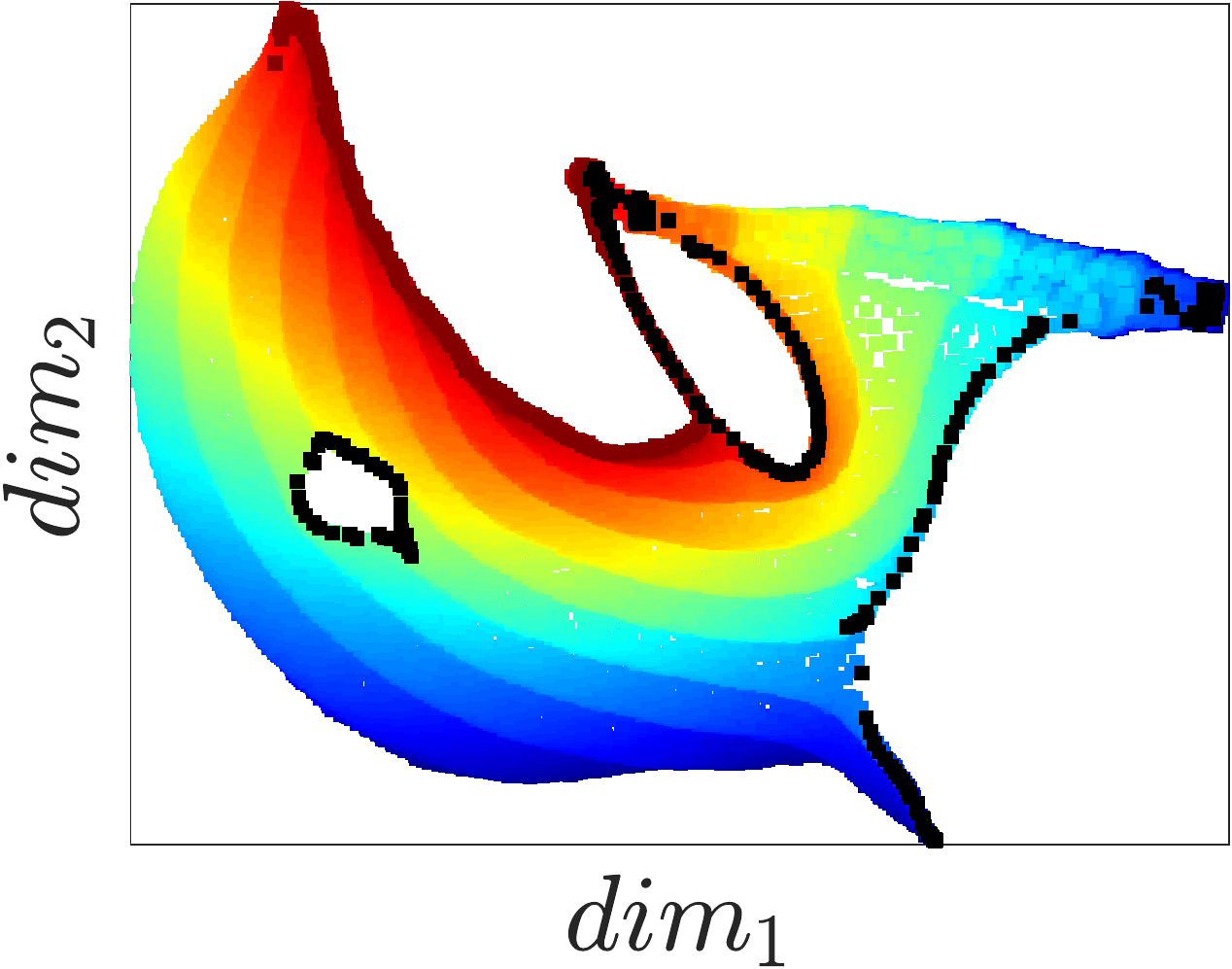}}  
    \caption{Multiple CHs: 2D embedding of different algorithms.}
   \label{fig:CH2}
\end{figure}

{The UMAP is also amiable to capture multiple holes of different sizes. An example with 3 CHs in the SA is shown in~Fig.~\ref{fig:CH2}(a). Again, PCA,  t-SNE, and SE fail to capture the visual holes properly, while UMAP and LLE capture them smoothly. However, the LLE provides several small visual holes that can be interpreted as CHs. When a CH is not enclosed (the bottom CH in Fig.~\ref{fig:CH2}(a)), then there is no enclosed hole in the embedding. In that case, there will be anomalies such as a long-tail indicating a possible CH nearby. Once these visual CH boundaries are identified from the UMAP embedding (manually or automated via  boundary detection), then the localization techniques~\cite{ruble2018wireless} can be applied to the selected channel samples from the boundary points to obtain the true-geographical CH boundary locations.}\looseness = -1

\vspace{-.3cm}
\section{Conclusion}\label{Sec:Conclusion}
In this work, we demonstrated first time in the literature that UMAP can reliably detect CHs from unlabelled UE channel samples available at the BSs. The CHs appear as visual holes in the low-dimensional embeddings due to the preservation of the neighborhood structure inherent in the channel samples without using any measurement reports or location information from the UEs. {Building on this proposed framework, our future work includes assessment of the detection probability of CHs in line-of-sight (LOS) and non-LOS scenarios and the detection of traffic holes with the noisy channel estimates.}\looseness = -1 
\vspace{-.2cm}

\vspace{-.15cm}
\bibliographystyle{IEEEtran}
\bibliography{IEEEabrv,references}

\begin{thebibliography}{10}
\providecommand{\url}[1]{#1}
\csname url@samestyle\endcsname
\providecommand{\newblock}{\relax}
\providecommand{\bibinfo}[2]{#2}
\providecommand{\BIBentrySTDinterwordspacing}{\spaceskip=0pt\relax}
\providecommand{\BIBentryALTinterwordstretchfactor}{4}
\providecommand{\BIBentryALTinterwordspacing}{\spaceskip=\fontdimen2\font plus
\BIBentryALTinterwordstretchfactor\fontdimen3\font minus
  \fontdimen4\font\relax}
\providecommand{\BIBforeignlanguage}[2]{{%
\expandafter\ifx\csname l@#1\endcsname\relax
\typeout{** WARNING: IEEEtran.bst: No hyphenation pattern has been}%
\typeout{** loaded for the language `#1'. Using the pattern for}%
\typeout{** the default language instead.}%
\else
\language=\csname l@#1\endcsname
\fi
#2}}
\providecommand{\BIBdecl}{\relax}
\BIBdecl

\bibitem{gomez2016method}
A.~G{\'o}mez-Andrades, R.~Barco, and I.~Serrano, ``A method of assessment of
  {LTE} coverage holes,'' \emph{EURASIP J. Wireless Commun. Netw.}, vol. 2016,
  no.~1, pp. 1--12, Dec 2016.

\bibitem{anjinappa2020base}
C.~K. Anjinappa, F.~Erden, and I.~Güvenç, ``Base station and passive
  reflectors placement for urban mmwave networks,'' \emph{{IEEE} Trans.\ Veh.\
  Technol.}, vol.~70, no.~4, pp. 3525--3539, Apr 2021.

\bibitem{akbari2016reliable}
I.~Akbari, O.~Onireti, A.~Imran, M.~A. Imran, and R.~Tafazolli, ``How reliable
  is {MDT}-based autonomous coverage estimation in the presence of user and
  {BS} positioning error?'' \emph{{IEEE} Wireless Commun. Lett.}, vol.~5,
  no.~2, pp. 196--199, Jan 2016.

\bibitem{hapsari2012minimization}
W.~A. Hapsari, A.~Umesh, M.~Iwamura, M.~Tomala, B.~Gyula, and B.~Sebire,
  ``Minimization of drive tests solution in {3GPP},'' \emph{{IEEE} Commun.
  Mag.}, vol.~50, no.~6, pp. 28--36, June 2012.

\bibitem{grissa2017location}
M.~Grissa, B.~Hamdaoui, and A.~A. Yavuza, ``Location privacy in cognitive radio
  networks: {A} survey,'' \emph{IEEE Commun. Surveys Tuts.}, vol.~19, no.~3,
  pp. 1726--1760, Apr 2017.

\bibitem{ruble2018wireless}
M.~Ruble and I.~G{\"u}ven{\c{c}}, ``Wireless localization for mmwave networks
  in urban environments,'' \emph{EURASIP J. Adv. Sig. Proc.}, vol. 2018, no.~1,
  pp. 1--19, Dec 2018.

\bibitem{mcinnes2018umap}
L.~McInnes, J.~Healy, and J.~Melville, ``{UMAP}: Uniform manifold approximation
  and projection for dimension reduction,'' \emph{arXiv preprint
  arXiv:1802.03426}, Feb 2018.

\bibitem{studer2018channel}
C.~Studer, S.~Medjkouh, E.~G{\"o}n{\"u}lta{\c{s}}, T.~Goldstein, and
  O.~Tirkkonen, ``Channel charting: Locating users within the radio environment
  using~channel state information,'' \emph{IEEE Access}, vol.~6, pp.
  47\,682--47\,698, Aug ~2018.

\bibitem{Virtual_Rep}
W.~U. {Bajwa}, J.~{Haupt}, A.~M. {Sayeed}, and R.~{Nowak}, ``Compressed channel
  sensing: A new approach to estimating sparse multipath channels,''
  \emph{Proc. {IEEE}}, vol.~98, no.~6, pp. 1058--1076, Apr 2010.

\bibitem{Alkhateeb2019}
A.~Alkhateeb, ``{DeepMIMO}: A generic deep learning dataset for millimeter wave
  and massive {MIMO} applications,'' in \emph{Proc. Inf. Theory Appl. Wkshp.
  (ITA)}, San Diego, CA, Feb 2019, pp. 1--8.

\bibitem{pedregosa2011scikit}
F.~Pedregosa, G.~Varoquaux, A.~Gramfort, V.~Michel, B.~Thirion, O.~Grisel,
  M.~Blondel, P.~Prettenhofer, R.~Weiss, V.~Dubourg \emph{et~al.},
  ``Scikit-learn: Machine learning in python,'' \emph{J. Mach. Learn. Res.},
  vol.~12, pp. 2825--2830, Nov 2011.

\bibitem{venna2005local}
J.~Venna and S.~Kaski, ``Local multidimensional scaling with controlled
  tradeoff between trustworthiness and continuity,'' in \emph{Wkshp. on
  Self-Organizing Maps}.\hskip 1em plus 0.5em minus 0.4em\relax Citeseer, Sep
  2005, pp. 695--702.

\end{thebibliography}



\end{document}